\documentclass[10pt,twocolumn,letterpaper]{article}

\usepackage[pagenumbers]{cvpr} 

\usepackage{graphicx}
\usepackage{amsmath}
\usepackage{amssymb}
\usepackage{booktabs}
\usepackage{bm}
\usepackage{bbm}
\usepackage[boxed]{algorithm2e}
\usepackage{multirow}
\usepackage[accsupp]{axessibility}  

\usepackage{ragged2e}
\usepackage[pagebackref,breaklinks,colorlinks]{hyperref}

\usepackage[capitalize]{cleveref}
\crefname{section}{Sec.}{Secs.}
\Crefname{section}{Section}{Sections}
\Crefname{table}{Table}{Tables}
\crefname{table}{Tab.}{Tabs.}

\graphicspath{{figures/}}

\def\cvprPaperID{8168} 
\def\confName{CVPR}
\def\confYear{2022}

\DeclareMathOperator{\E}{\mathbb{E}}
\newcommand{\x}{\bm{x}}
\newcommand{\e}{\bm{e}}
\newcommand{\ux}{\bm{u}}

\newcommand{\z}{\bm{z}}

\newcommand{\y}{\bm{y}}

\newcommand{\I}{\bm{I}}
\newcommand{\s}{\bm{s}}
\newcommand{\1}{\mathbbm{1}}
\newcommand{\zerov}{\bm{0}}

\newcommand{\real}{\mathbb{R}}
\newcommand{\loss}{\mathcal{L}}
\newcommand{\Dloss}{\loss_{D}}
\newcommand{\Gloss}{\loss_{G}}
\newcommand{\lblloss}{\loss_{\textrm{adv}}^{\textrm{lbl}}}
\newcommand{\unlblloss}{\loss_{\textrm{adv}}^{\textrm{unlbl}}}
\newcommand{\clsloss}{\loss_{\textrm{cls}}}

\DeclareMathOperator*{\argmax}{arg\,max}

\newcommand{\pxy}{(\x, \y)\sim p(\x, \y)}
\newcommand{\pu}{\ux\sim p(\ux)}
\newcommand{\qzy}{(\z, \y)\sim q(\z, \y)}  
\newcommand{\Dparam}{\theta_D}
\newcommand{\Gparam}{\theta_G}
\newcommand{\Cparam}{\theta_C}
\DeclareMathOperator{\Adam}{AdamOptimizer}

\begin{document}

\title{OSSGAN: Open-Set Semi-Supervised Image Generation}

\author{Kai Katsumata \qquad Duc Minh Vo \qquad Hideki Nakayama\\
The University of Tokyo, Japan\\
{\tt\small \{katsumata,vmduc,nakayama\}@nlab.ci.i.u-tokyo.ac.jp}
}

\captionsetup[sub]{justification=centering}

\twocolumn[{%
\renewcommand\twocolumn[1][]{#1}%
\maketitle
\begin{center}
    \centering
    \captionsetup{type=figure}
    \begin{minipage}[t]{0.333\textwidth}
    \centering
      \includegraphics[width=0.98\textwidth]{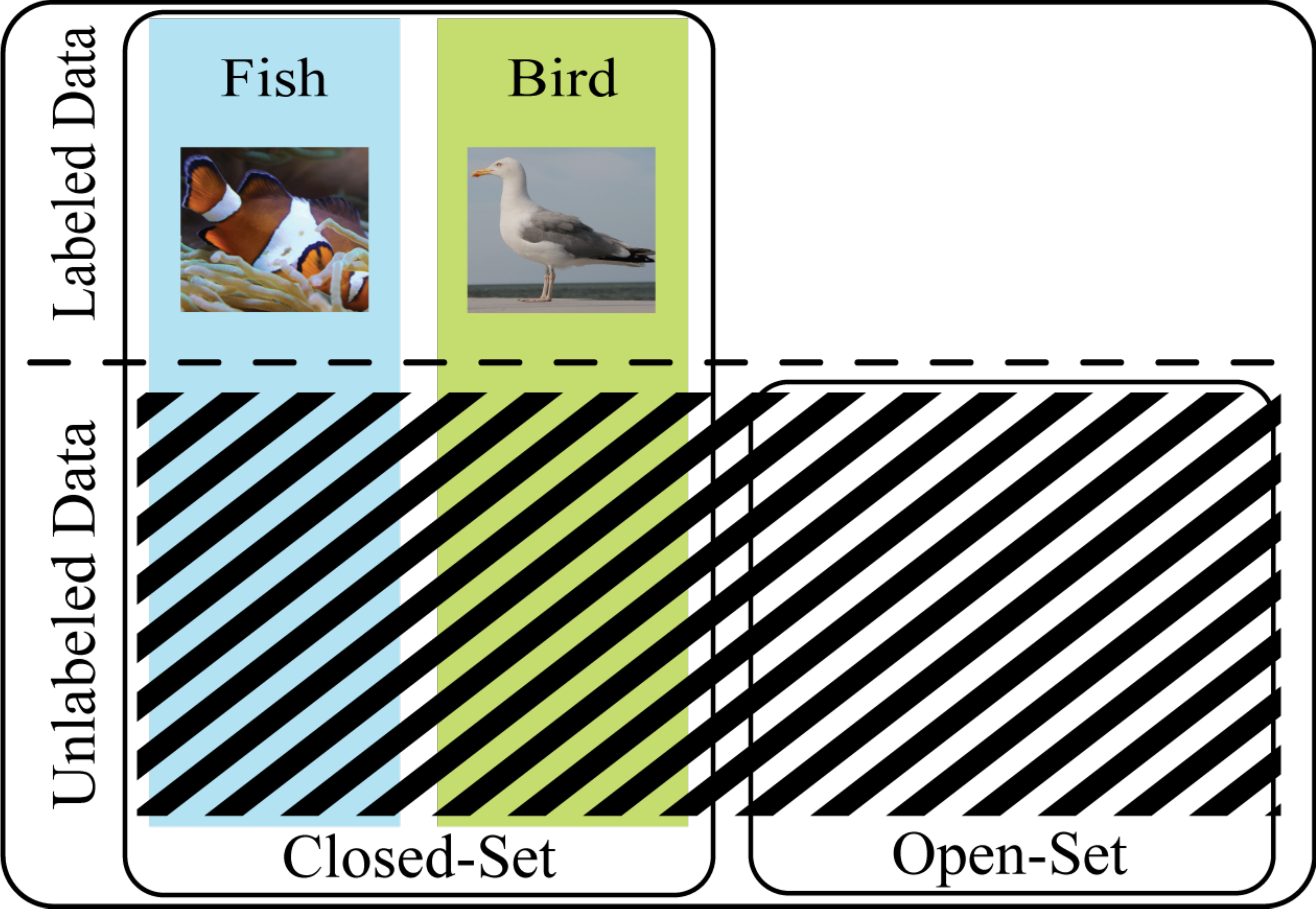} 
      \subcaption{\footnotesize Conventional image generation} \label{fig:cgan}
    \end{minipage}%
    \begin{minipage}[t]{0.333\textwidth}
    \centering
      \includegraphics[width=0.98\textwidth]{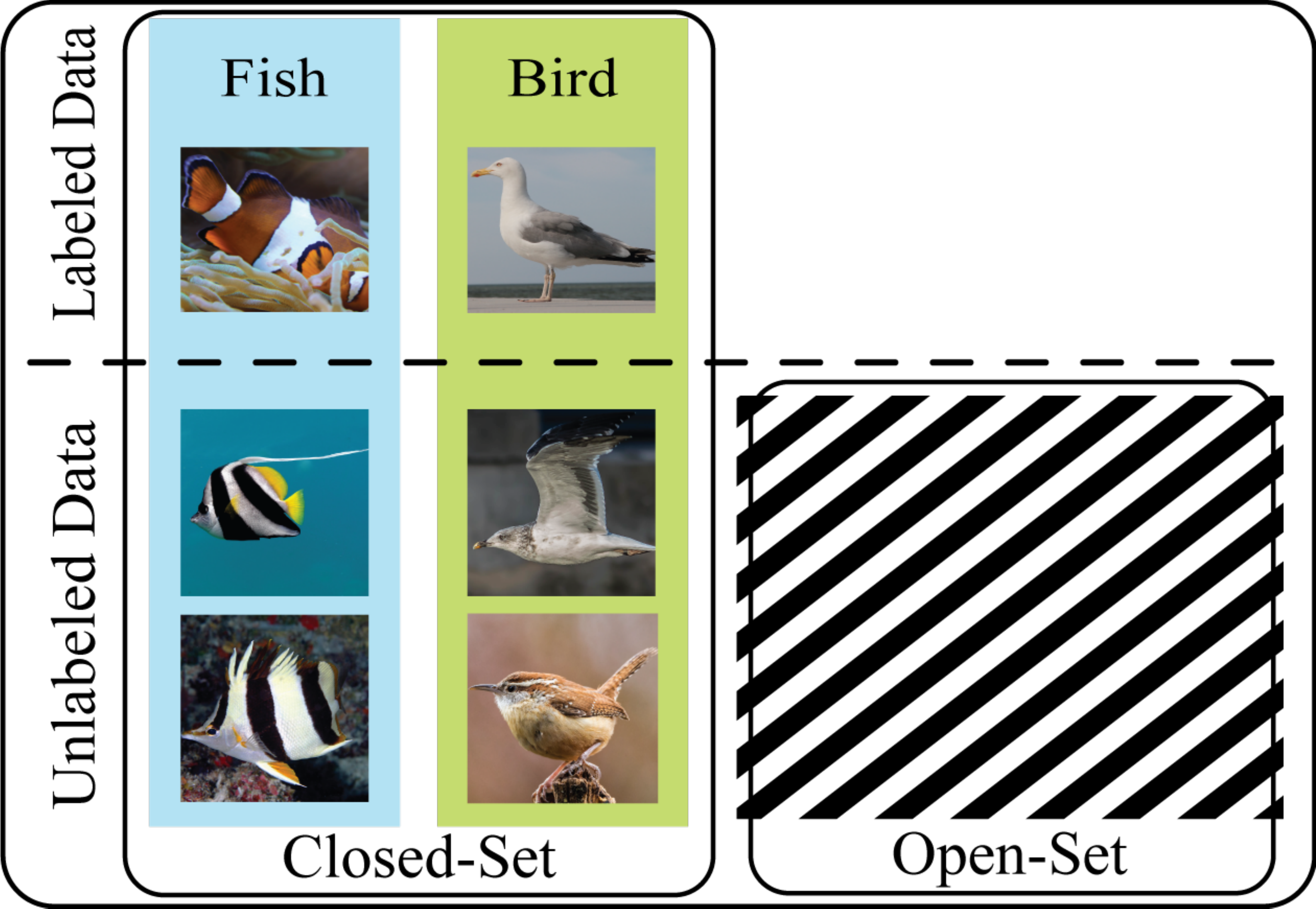}
      \subcaption{\footnotesize (Closed-set) semi-supervised image generation}\label{fig:ssgan}
    \end{minipage}%
    \begin{minipage}[t]{0.333\textwidth}
    \centering
      \includegraphics[width=0.98\textwidth]{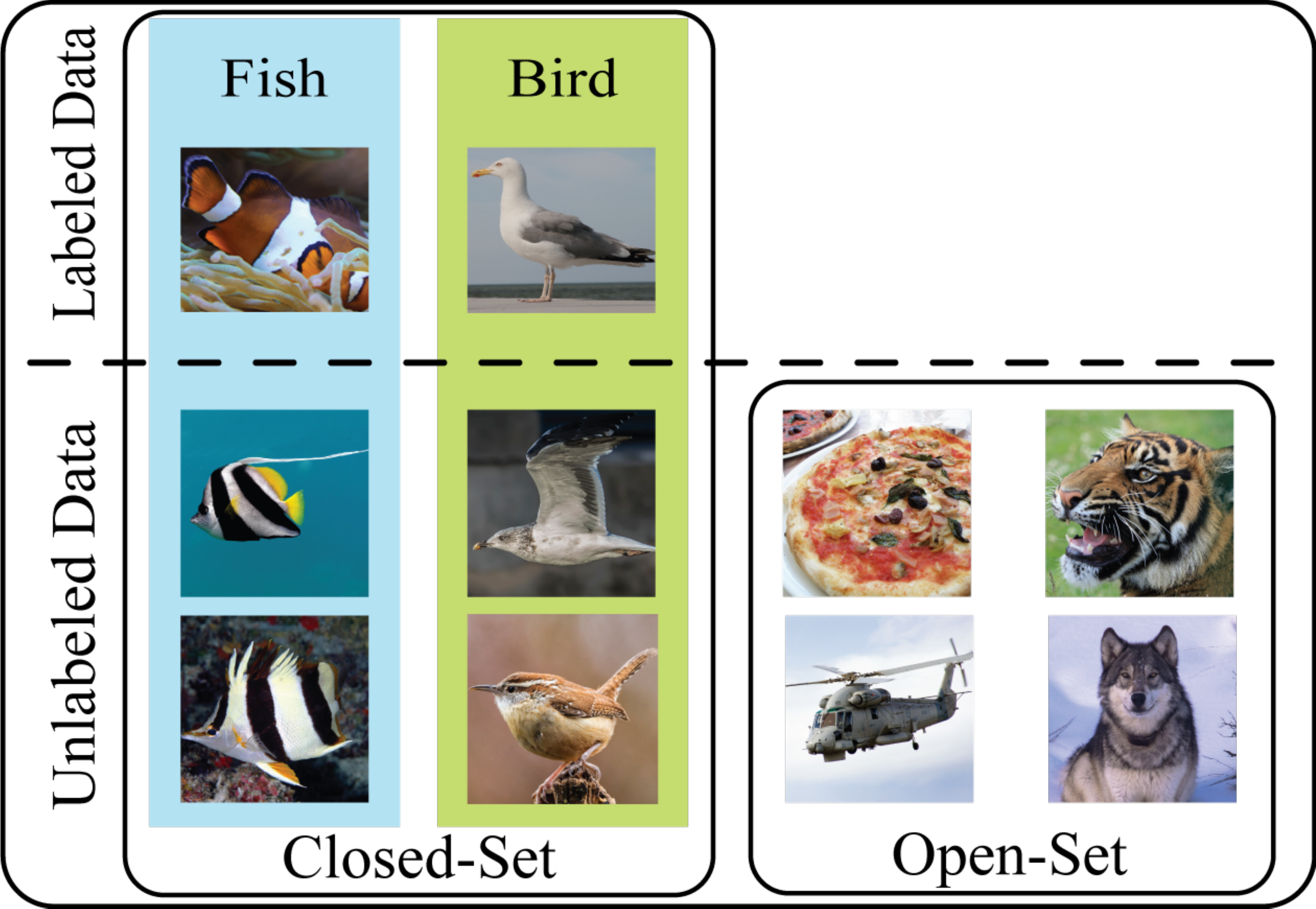}
      \subcaption{\footnotesize Open-set semi-supervised image generation \\ (our task)} \label{fig:osgan}
    \end{minipage}%
    \caption{We explore the conditional image generation in which we relax the assumption about training data. 
    (a) Conventional supervised
      image generation~\cite{Mirza2014,Zhang2019,Brock2018} principally uses labeled data for training. (b) (Closed-set) semi-supervised image generation~\cite{Lucic2019} generalizes the supervised image generation, allowing the presence of unlabeled data with
      samples inside classes of interest, which are closed-set samples.
      (c) Our task of open-set semi-supervised image generation further generalizes
      semi-supervised image generation. The unlabeled data
      contain closed-set samples and open-set samples, which do not belong to any of the classes of
      interest.} \label{fig:task}
\end{center}%
}]

\begin{abstract}
We introduce a challenging training scheme of conditional GANs, called open-set semi-supervised image generation, where the training dataset consists of two parts: (i) labeled data and (ii) unlabeled data with samples belonging to one of the labeled data classes, namely, a closed-set, and samples not belonging to any of the labeled data classes, namely, an open-set.
Unlike the existing semi-supervised image generation task, where unlabeled data only contain closed-set samples, our task is more general and lowers the data collection cost in practice by allowing open-set samples to appear.
Thanks to entropy regularization, the classifier that is trained on labeled data is able to quantify sample-wise importance to the training of cGAN as confidence, allowing us to use all samples in unlabeled data.
We design OSSGAN, which provides decision clues to the discriminator on the basis of whether an unlabeled image belongs to one or none of the classes of interest, 
smoothly integrating labeled and unlabeled data during training.
The results of experiments on Tiny ImageNet and ImageNet show notable improvements over supervised BigGAN and semi-supervised methods.
Our code is available at \url{https://github.com/raven38/OSSGAN}.
\end{abstract}

\section{Introduction}
\label{sec:intro}

The outstanding performance of the SoTA conditional generative adversarial networks (cGANs)~\cite{Mirza2014,Zhang2019,Brock2018} is heavily reliant on having access to a vast amount of labeled data during training (\cref{fig:cgan}).
This dependence necessitates significant efforts to label the data and limits the applications of cGANs in real-world scenarios.
Reducing the reliance on labeled data in training cGANs is thus deemed necessary.

Semi-supervised image generation~\cite{Lucic2019,Sricharan2017,Springenberg2016,Li2017,Deng2017} allows the appearance of both labeled and unlabeled data during training, with the unlabeled data primarily containing within classes of interest (closed-set samples) (\cref{fig:ssgan}).
Despite the advances, the unlabeled data assumption is at odds with the fact that the majority of unlabeled data is outside of classes of interest (open-set samples), and ensuring that unlabeled data do not contain open-set samples is often costly and prone to error.
In fact, in ~\cite{Lucic2019,Sricharan2017,Springenberg2016,Li2017,Deng2017}, even open-set samples are classified into classes that appear in labeled data, resulting in cGAN performance deterioration.

We go beyond semi-supervised image generation by allowing the use of unlabeled data gathered miscellaneously to reduce the effort of labeling and introduce a novel task of open-set semi-supervised image generation (\cref{fig:osgan}).
Unlabeled data contain open-set samples, and the conditional generator should produce images that are indistinguishable from real ones even when trained on both labeled and unlabeled data.
Unlike the conventional semi-supervised fashion, the task allows unlabeled data to have the category set mismatched with labeled data, reducing the required labeling effort.
Our task is a significant step towards real-world data, which contains labeled data and unlabeled data (both closed-set and open-set samples), lowering the data construction cost and expanding the range of real-world applications of cGANs.

To address our new task, we design \textbf{O}pen-\textbf{S}et \textbf{S}emi-supervised GAN (OSSGAN).
We simultaneously train cGAN and a classifier that assigns labels to unlabeled data.
By incorporating entropy regularization into the cross-entropy loss,
the classifier quantifies the confidence of the prediction
to enable the discriminator to use unlabeled data, including open-set samples, smoothly. Consequently, OSSGAN allows
the natural integration of unlabeled data into cGANs without explicitly eliminating open-set samples.

The results of empirical experiments demonstrate that OSSGAN effectively utilizes
unlabeled data, including open-set samples. More importantly, we achieve better performance in terms of FID and other metrics
against strong supervised and semi-supervised baselines.  
Notably, our
method achieves a performance comparable to that of BigGAN~\cite{Brock2018}, which has up to five times as many
labeled samples as OSSGAN.  Furthermore, the experiments with different
degrees of an open-set sample ratio show that the proposed method is robust to
miscellaneous data.  
Qualitative experimental results also
reveal the superiority of our method.
Our contributions are summarized as follows.
\begin{itemize}
\item We propose a novel open-set semi-supervised image generation task, which is based on a relaxed assumption in the case of building a dataset at a reasonable cost. 
\item We design OSSGAN, thanks to entropy regularization, smoothly using closed- and open-set samples in unlabeled data in cGAN training. 
\item We demonstrate the superiority of the proposed method over baselines on several benchmarks with limited labeled data in terms of quantitative metrics such as FID. Our qualitative experiments also show that our method achieves better generation quality.
 \end{itemize}

\section{Related work}
\label{sec:related}
\noindent \textbf{CGANs}~\cite{Mirza2014} are a GANs extension, which learns a
conditional generative distribution.  
CGANs can deal with many
types of conditions such as class label~\cite{Mirza2014}, text
description~\cite{Zhang2017}, or another image~\cite{Isola2017}.
For
well-constructed datasets,~\cite{Brock2018,Zhang2019,Miyato2018} are proposed to
improve quality, fidelity, and training stability.  Among them,
self-attention GAN~\cite{Zhang2019} and BigGAN~\cite{Brock2018} outperform other GANs with hundreds of classes. 
The progress in network architectures, optimization algorithms, and the quality and quantity of datasets support high-fidelity image generation.
We aim to achieve high-fidelity image generation without a well-constructed dataset. 

\noindent \textbf{Image generation with data constraints} is aimed at improving the generation quality without using enormous amounts of data.
Collecting a large labeled dataset requires a tremendous annotation
cost. To achieve better performance within limited resources
(\ie, time and money), several
studies~\cite{Shaham2019,Hinz2021,Zhao2020,Karras2020} contribute to
the data-efficient aspect of cGANs.  Another line of
studies~\cite{Ting2019,Lucic2019} focusing on the fact that unlabeled
images are easier to collect than labeled images employs
semi-supervised or unsupervised fashion.  For semi-supervised image
generation, ~\cite{Sricharan2017,Springenberg2016} employ an
unconditional discriminator for unlabeled data, and
~\cite{Li2017,Deng2017} take pseudo labels by employing a
classifier. Unlabeled data have been efficiently utilized in
self-supervised learning~\cite{Lucic2019}. A different aspect in the study of semi-supervised learning and GAN concerns semi-supervised recognition tasks that employ GAN for generating pseudo samples~\cite{Salimans2016,Dai2017}.
 Unsupervised image
generation frees us from tedious annotation labor.
However, unsupervised
methods do not control generated outcomes as semi-supervised methods
do. 
In this study, we utilize unlabeled data gathered
miscellaneously for training cGANs to reduce the data construction
cost, which will further broaden their range of application.

\noindent \textbf{Open-set semi-supervised recognition} has the same
objective as our method but addresses a totally different task.  The goal of
the recognition task is to build a model distinguishing open-set
samples using a dataset consisting of labeled data with only
closed-set classes and unlabeled data with both closed- and open-set
classes.  The joint optimization of classification and open-set sample
detection models~\cite{Yu2020} and the consistent regularization with
data augmentation~\cite{Luo2021} are used to tackle the problem.  In
contrast to recognition paradigms aimed at separating explicitly open-
and closed-set samples, our generation task does not necessarily
require explicit separation of these samples.
Instead of applying a method for detecting open-set
samples, we investigate a method of utilizing open-set samples.

\section{Open-set semi-supervised image generation}
\label{sec:task}

\subsection{Task definition}

While labeling data is tremendously costly, we can collect unlabeled data at a relatively low cost.
However, as the semi-supervised image generation task does not allow the unlabeled data to contain any open-set samples, filtering out such open-set samples is mandatory.
As a step towards reducing the manpower expended in the filtering process for building closed-set semi-supervised data, we consider a learning framework without the assumption that the sets of classes are shared between labeled and unlabeled data, meaning that open-set samples are freely applicable.
While the aim of our task is to generate the images with the classes of interest (labeled data) similarly to supervised and semi-supervised image generation~\cite{Brock2018,Zhang2019,Miyato2018,Lucic2019}, we assume the unlabeled data contain closed-set and open-set samples (\cref{fig:task}).

Let $\x, \ux \in \real^{d}$ be real labeled and unlabeled samples, respectively, where $d = H\times W\times 3$ is the dimension of a sample where $H$ and $W$ are the height and width of the image.
Let $y \in \mathcal{Y} = \{1,2,\ldots,K\}$ be the class where $K$ is the number of known classes.
For the sake of simplicity, we will use a probability vector $\y \in \{\y \in \real^{K} | \sum_{1}^{K}=1\ \textrm{and}\ 0 \leq \y_i\}$ 
when referring to the class unless otherwise specified.
Thus, class $y$ can be interpreted as a one-hot vector $\y = [\{\1[y=j]\}_{j=1}^{K}]^{\mathsf{T}}$.
We denote the distribution of labeled data as $p(\x, \y)$, whereas the distribution of unlabeled data is $p(\ux)$. 
The unlabeled data contain the samples that cannot be classified into one of the $K$ classes, and such samples are considered as open-set samples.

We are given a set of labeled data $\mathcal{D}_l = \{(\x_i, \y_i)\}^{n_l}_{i=1} \sim p(\x, \y)$ with $n_l$ samples and a set of unlabeled data $\mathcal{D}_u = \{\ux_i\}^{n_u}_{i=1} \sim p(\ux)$ with $n_u$ samples as training data. 
 By using the latent variable $\z \in \real^{l}$ with $l$ being the dimension of the latent variable, the prior distribution is $q(\z,\y) =
q(\z)q(\y)$ with a Gaussian distribution $q(\z)$ and a uniform
distribution $q(\y)$ on $\{\e^{(1)}, \e^{(2)},\ldots,\e^{(K-1)},\e^{(K)}\}$, where $\e^{(i)}$ is the $i$-th standard basis vector of $\real^K$.  The task of open-set semi-supervised image
generation is to learn a generator $G: \real^{l} \times \real^{K} \to \real^{d}$
such that $G$ takes a latent variable and a label $\y \sim q(\y)$
and generates images that are indistinguishable from real ones for
a given label.

\subsection{General training objective functions}

To accomplish the task we proposed, the discriminator should be able to deal with the unlabeled data (and the labeled data), that is, it should easily distinguish between closed-set samples and open-set samples.
A straightforward way to do this is to use an auxiliary classifier to assign a class to unlabeled samples.
Motivated by this observation, we define a cGAN with an auxiliary classifier as follows.
Given a generator $G$, a discriminator $D$, and an auxiliary classifier $C$, the general discriminator loss $\mathcal{L}_{D}$ and generator loss $\mathcal{L}_G$ are defined as
\begin{align}
\Dloss = & \lblloss + \unlblloss + \lambda \clsloss \\
\Gloss = & \E_{\qzy} [-D(G(\z, \y), \y)],\label{eq:gloss}
\end{align}
where $\lblloss$ and $\unlblloss$ are adversarial losses for labeled and unlabeled data, respectively.
The classifier loss $\clsloss$ has a hyperparameter $\lambda$ for balancing the loss terms.
The $\lblloss$ follows the conventional adversarial loss with hinge loss $f_{D}(\cdot) = \max(0, 1+\cdot)$ for labeled data:
\begin{align}
  \lblloss = & \E_{\pxy} [ f_{D}(-D(\x, \y))] \nonumber \\
                  & +\E_{\qzy} [ f_{D}(D(G(\z, \y), \y))].\label{eq:lblloss}
\end{align}
Naively, the classifier loss $\clsloss$ can be cross-entropy loss:
\begin{align}
  \clsloss =  & - \E_{\pxy} [\log p_{c}(\y^\mathsf{T}\s | \x)].
\end{align}
We note that the discriminator $D$ and the classifier $C$ share the feature extractor part.
Mathematically, $D: \real^{d} \times \real^{K} \to \real$ takes the form $D(\x, \y) = W_{1}\tilde{D}(\x) + b + \tilde{D}(\x)^\mathsf{T} W_{2} \y$,
$\tilde{D}: \real^{d} \times \real^{K} \to \real^{h}$ is a feature extractor in $D$,
$\s$ is
$[0, 1, 2,\ldots,K-1]^\textsf{T}$, and
$p_{c}(y | \x) = C_y(\tilde{D}(\x))$ is the $y$-th output of the auxiliary classifier $C: \real^{h} \to \real^{K}$.
Here, $W_{1} \in \real^{h \times 1}$ is the weight parameter of the
fully connected layer, $W_{2} \in \real^{h \times K}$ is the embedding
matrix for a label, and $h$ is the dimension of the extracted feature.
The scalar representation of a label is represented by $\y^\mathsf{T}\s$ with a one-hot vector $\y$.
We will further adaptively customize $\lblloss$, $\unlblloss$, and $\clsloss$ for different methods, as discussed later.

\section{Proposed method}
\label{sec:method}

\subsection{Threshold-based method}
\label{sec:threshold_method}

We introduce two baseline methods for our task, called RejectGAN and OpensetGAN.
These methods are semi-supervised GANs extended by employing a classifier to assign predicted classes to unlabeled samples as new labels.
RejectGAN only considers labeled samples and closed-set samples in unlabeled data in the training of GANs by filtering out unlabeled samples that have low confidence, as open-set samples.
OpensetGAN explicitly utilizes open-set samples in the unlabeled data by assigning $K\!+\!1$ classes, the open-set class, to unlabeled samples with low confidence.
Below, we describe the details of these methods.

\noindent \textbf{RejectGAN}.
The principal purpose of this method is to train cGANs on the dataset with sufficient volume and clean labels. 
Whenever the labels of samples are available, we use them to update an auxiliary classifier in the same manner as ACGAN~\cite{Odena2017}.
At the same time, we assign the predicted classes to unlabeled samples when the classifier predictions have a probability associated with a predicted class equal to or higher than a threshold.
Then, by eliminating unlabeled samples with a probability associated with the predicated class less than a threshold, we train the discriminator only with unlabeled samples with high confidence, labeled samples, and samples synthesized by a generator.
Accordingly, we modify the adversarial loss for unlabeled samples as
\begin{align}
\unlblloss = \E_{\pu} [f_{D}(-D(\ux, \hat{\y})) | \max(\tilde{\y}) \geq c],
\end{align}
where $c$ is a threshold of confidence, $\tilde{\y} = C(\tilde{D}(\ux))$ is a predicted probability vector, and $\hat{\y} = \e^{(\argmax\tilde{\y})}$ is a predicted label with $\e^{(i)}$ being the $i$-th
standard basis vector of $\real^{K}$. 
A confidence score larger than the threshold
means that the sample is clearly classified into $y \in \mathcal{Y}$. We ignore the samples with a confidence score lower than the threshold in calculating the loss.

\noindent \textbf{OpensetGAN}. 
In addition to learning class-specific features, this method is also aimed at learning class-invariant image features by adding a novel class for the open-set samples to the discriminator inputs.
In contrast to RejectGAN, the discriminator $D: \real^{d} \times \real^{K+1} \to \real$ takes a sample and a label vector with the length of $K+1$ to consider open-set samples detected by the auxiliary classifier.
Accordingly, OpensetGAN extends the adversarial losses $\lblloss$ and $\Gloss$ to accept the $K+1$-dimensional condition vector:
\begin{align}
  \lblloss =  & \E_{\pxy} [ f_{D}(-D(\x, \y^\mathsf{T}\!\I_{K,K\!+\!1}))] \nonumber \\
              & +\E_{\qzy} [ f_{D}(D(G(\z, \y), \y^\mathsf{T}\!\I_{K,K\!+\!1}))],\\
  \mathcal{L}_G = & \E_{\qzy} [-D(G(\z, \y), \y^\mathsf{T}\!\I_{K,K\!+\!1})],
\end{align}
where $\I_{K,K\!+\!1} = (\I_{K}\ \zerov)$ is the $K\!\times\!(K\!+\!1)$ rectangular identity matrix with the $K\!\times\!K$ identity matrix $\I_{K}$ and the zero vector $\zerov$. 
OpensetGAN assigns the $K\!+\!1$ classes to unlabeled data to train on the labeled data with the $K$ classes, and then the adversarial loss for unlabeled samples is
\begin{align}
  \unlblloss =  & \E_{\pu} [f_{D}(-D(\ux, \hat{\y}))].
\end{align}
Here, we define 
the classifier output $\hat{\y}\in\real^{K+1}$ as
\begin{align}
  \hat{\y} = 
  \begin{cases}
    \e^{(\argmax\tilde{\y})}, & \text{if } \max(\tilde{\y}) \geq c \\
    \e^{(K+1)}, & \text{otherwise},
  \end{cases}\label{eq:distribute_function}
\end{align}
and we denote the $K$ known classes and an additional open-set class by $\{\e^{(1)}, \e^{(2)},\ldots,\e^{(K+1)}\}$ with the standard basis vectors of $\real^{K+1}$.
The convert matrix $\I_{K,K+1}$ adds the zero-filled $K\!+\!1$-th column to a
label vector for a discriminator.  In \cref{eq:distribute_function}, we
assign one of the known classes to unlabeled samples with
high confidence score and the open-set class to unlabeled samples
with low confidence score.

\subsection{The intuition behind OSSGAN}
\label{sec:idea}

\begin{figure}[tb]
\centering
\includegraphics[width=.95\linewidth]{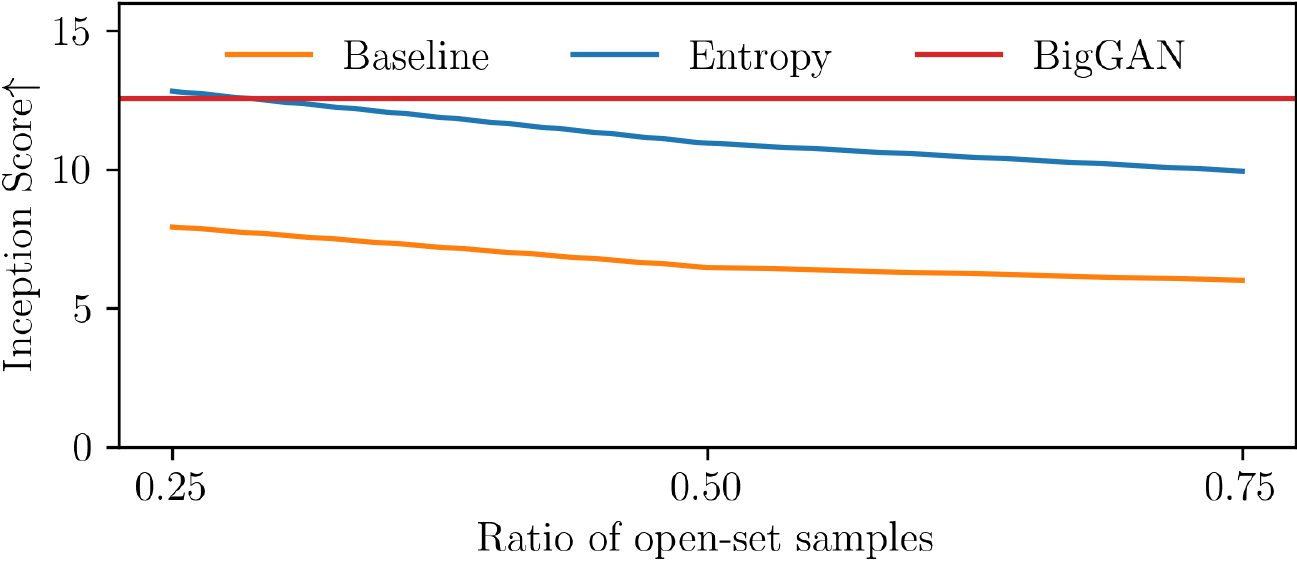}
\caption{Effect of assigning high entropy labels to open-set
  samples on Tiny ImageNet. Entropy indicates the method of verifying our approach. BigGAN is trained on only labeled
  samples. Here, the experiments contain enough labeled samples for
  training cGANs.  The method assigning high entropy labels to open-set
  samples maintains the performance of BigGAN.
}\label{fig:entropy}
\end{figure}

The methods described above do not efficiently exploit a given dataset because employing a threshold hinders the training of GAN.
Furthermore, their performance is sensitive to the threshold, and the average entropy of predicted probabilities shifts with each iteration.
In the training phase, finding the optimal threshold in each iteration is difficult because we do not know the ratio of known to unknown samples in unlabeled samples. 
As a result, because it eliminates closed-set samples with low confidence score in unlabeled data, RejectGAN misses out on useful information for training a generator and discriminator.
Similarly, OpensetGAN may fail to learn the class feature owing to misclassifying known class samples as open-set samples.
Since these flaws cause the instability of threshold-based methods, we require a more robust method that does not need careful tuning.

To devise a threshold-free method, we use the entropy of the label as a confidence score and feed continuous labels with confidence into the discriminator. Then, the discriminator uses the information that samples with low entropy are clearly classified into the known classes, and samples with high entropy are not classified into the known classes, resulting in no unlabeled samples being missed.

 To ensure that the idea of assigning high entropy labels to open-set samples is effective for the task, we compare the method of manually assigning high entropy labels (as an oracle) to open-set samples with supervised and semi-supervised methods. 
The supervised method is BigGAN~\cite{Brock2018}. 
The semi-supervised method, which is referred to as Baseline in \cref{fig:entropy},  assigns one of the known classes to open-set samples with an auxiliary classifier.
BigGAN is only trained on closed-set samples. 
The high entropy method, which is abbreviated as Entropy, is trained on closed-set and open-set samples labeled $[1/K, 1/K, 1/K, \ldots,1/K]^\textsf{T}$. 
Entropy achieves a performance comparable to that of BigGAN in the case of a ratio of open-set samples of 25\%, as shown in \cref{fig:entropy}. 
It outperforms the baseline method in all other cases. 
These findings indicate that the strategy of assigning high entropy labels to open-set class samples helps to save training cGANs from contamination by open-set samples.

\subsection{OSSGAN}
\label{sec:osssgan}
\begin{figure}[tb]
\centering
\includegraphics[width=1.0\linewidth]{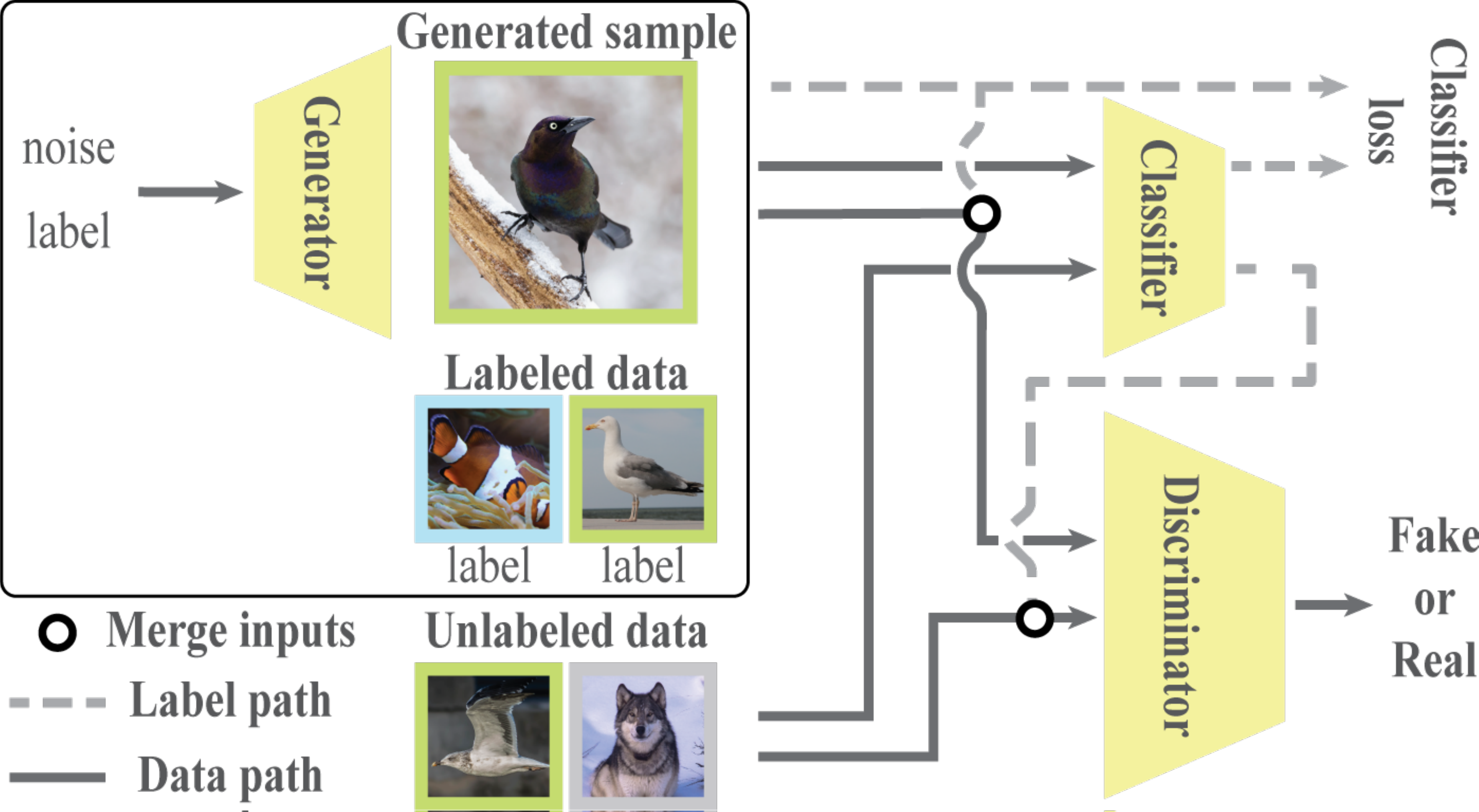}
\caption{Overview of the proposed method. A classifier is trained
  on labeled and generated data with the classifier loss consisting of cross-entropy and entropy regularization terms and infers labels for
  unlabeled data. The generator is trained in the same fashion as
  conventional cGANs. For the generated and labeled data, the
  discriminator is trained in the same way as conventional
  cGANs. It further takes unlabeled samples by regarding them as
  labeled samples. }\label{fig:method}
\end{figure}

To automatically assign high entropy labels to open-set samples in unlabeled data, we propose a method that quantifies the likelihood that a sample belongs to
any one of the closed-set classes and feeds the samples with likelihoods into the discriminator.
OSSGAN is made up of three parts: a generator, a discriminator, and an auxiliary classifier (\cref{fig:method}).
To train the auxiliary classifier, we use both real labeled and generated samples.
This is due to the fact that if the number of training data is insufficient, the auxiliary classifier performs poorly.
To train the discriminator, we use real labeled, real unlabeled, and generated samples.
While either the label of the real labeled samples or the label of the generated samples can be fed directly into the discriminator, we use the classifier output as the label of the unlabeled samples.
As a result, when the samples are labeled or generated, the discriminator $D\!:\!\real^{d}\!\times\!\real^{K}\!\!\to\real$ is fed a one-hot vector; when the samples are unlabeled, it is fed a continuous vector.
Our adversarial loss for unlabeled samples is defined as
\begin{align}
\unlblloss = \E_{\pu} [f_{D}(-D(\ux, C(\tilde{D}(\ux))))]\label{eq:ossgan_unlblloss}.
\end{align}

To provide more informative clues for identifying open-set samples from closed-set ones to the discriminator,
we introduce the entropy regularization term into the cross-entropy loss. 
While the standard classification loss, cross-entropy loss, results in low entropy, entropy regularization maintains high entropy of open-set samples in unlabeled data.
With the term maximizing the average entropy of the auxiliary classifier outputs, 
we customize $\clsloss$:
\begin{align}
  \clsloss = & - \E_{\pxy} [\log p_{c}(\y^\mathsf{T}\s | \x))]\nonumber \\
                  & - \E_{\pxy} [H(C(\tilde{D}(\x)))]\nonumber \\
                  & - \E_{\qzy} [\log p_{c}(\y^\mathsf{T}\s | G(\z, \y)))]\nonumber \\
                  & - \E_{\qzy} [H(C(\tilde{D}(G(\z, \y))))].\label{eq:ossgan_clsloss}
\end{align}
Without the regularization, the classifier predicts low entropy
values to all unlabeled samples, resulting in assigning a known
class even to unlabeled samples that should be treated as an open set.
The entropy term is defined by
  \begin{align}
    H(\x) = - \frac{\Sigma_{x \in \x} x \log x}{\log K}.
  \end{align}
  The range of the function is $0 \leq H(\x) \leq 1$.  We use
  a normalized function because it is difficult
  to set the hyperparameter $\lambda$ in the loss function when a
  function with a range of $0 \leq H(\x)$ can take a large value.  The contribution of the
  entropy regularization term varies with different $K$ in the experiment if we use the original
  entropy function.  
While the cross-entropy loss makes the entropy smaller, the entropy regularization makes the entropy larger. The cross-entropy loss more strongly affects closed-set samples, resulting in the clear separation between the closed- and open-set samples.
The number of labeled samples in the minibatch
  is quite small because unlabeled samples make up the majority of the
  dataset.  
To avoid a classifier that considers only the generated samples, which dominate the minibatch, we balance the ratio of the classification loss terms for labeled and generated samples by the number of labeled samples.
Finally, the overall objective function of OSSGAN is $\Gloss$ of \cref{eq:gloss} and $\Dloss$ consisting of \cref{eq:lblloss,eq:ossgan_unlblloss,eq:ossgan_clsloss}.

This method employs the raw probability vector for handling open-set
samples instead of using thresholds and is free from investigating the
optimal threshold.  
It also makes use of the inter-class similarity and class-invariant visual attributes that an auxiliary classifier acquires throughout the training process. 
For example, if the dog, cat, and monkey classes are given as known classes and a cow image is included in the open-set sample,
the image can be used as an image with common
mammalian attributes.

\subsection{Implementation details}
\label{sec:impl}

We choose BigGAN~\cite{Brock2018} as an example to verify OSSGAN.
Here, our method can be applied to different cGANs (\eg, SAGAN~\cite{Zhang2019} and SNGAN~\cite{Miyato2018}).
In detail, we build all the methods that are used in the experiments upon BigGAN~\cite{Brock2018} by
integrating DiffAugment~\cite{Zhao2020}.

For the experiments, we use the
hierarchical latent space with 20 dimensions for each latent variable
and the shared embedding with $d_{\z} = 128$.  We use minibatch sizes
of $1024$ and $256$ for the resolutions of $64 \times 64$ and $128 \times 128$, respectively. The learning rates are $1\times 10^{-4}$ and $4\times
10^{-4}$ for the generator and discriminator, respectively.

\section{Experiments setting}
\label{sec:experiment}

\begin{figure}[tb]
  \centering
\includegraphics[width=1.0\columnwidth]{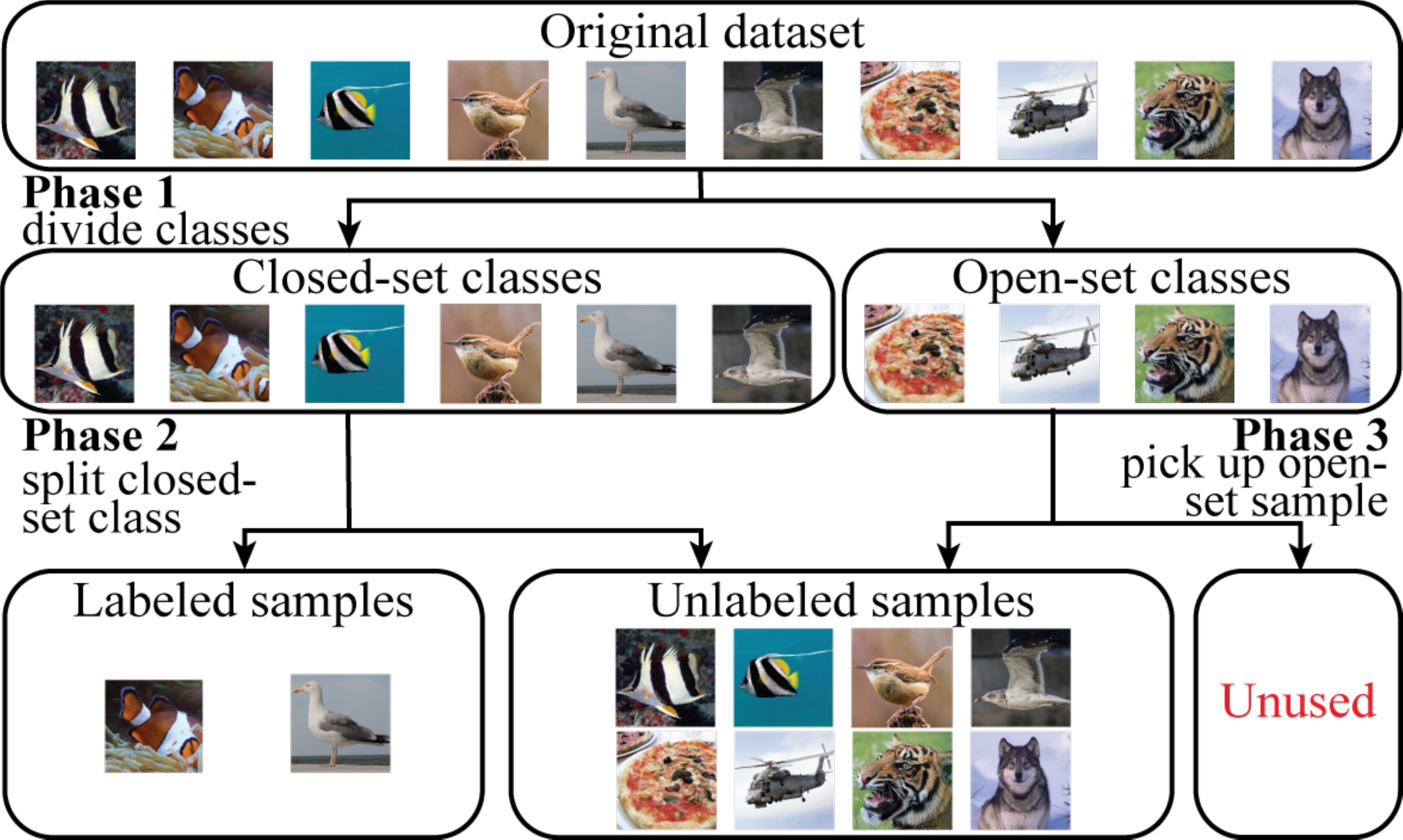}
\caption{Procedure of building open-set semi-supervised datasets. 
The procedure has three constants for governing the number of labeled, closed-set unlabeled, and open-set unlabeled samples.
} \label{fig:dataset}
\end{figure}

\noindent
\textbf{Datasets}. 
Using existing entirely labeled datasets, we create partially labeled datasets for benchmarking open-set semi-supervised image generation.
The dataset construction procedure is divided into three stages (\cref{fig:dataset}).
We use three constants: the number of closed-set classes, the ratio of labeled samples in closed-set class samples, and the usage ratio in open-set samples.
First, we divide the entirely labeled dataset into closed-set classes and open-set classes based on the number of closed-set classes.
Then, using the ratio of labeled samples in closed-set class samples, we split the closed-set samples into labeled and unlabeled samples.
Finally, we select samples from open-set samples based on the usage ratio in open-set samples, and we merge unlabeled samples in closed-set samples with the samples selected from open-set class samples. 

We use the Tiny ImageNet~\cite{TinyImagenet}, which consists of
200 diverse categories. Each class contains 500 and 100 images
for training and testing, respectively.  We use the
number of closed-set classes of $\{150, 100, 50\}$, the ratio of the
labeled samples in closed-set class samples of $\{0.95, 0.5, 0.2,
0.1\}$, and the usage ratio in open-set class samples of $1$. 
Then, using the above constants, we create 12 data configurations.
The easiest case has labeled samples with approximately
three-quarters of the dataset, and the hardest case has labeled
samples with less than 3\% of the dataset.

We also use ImageNet ILSVRC2012~\cite{Olga2015}.
It consists of $1000$ categories and provides $1.2$ million images.
For the dataset, we use the number of closed-set classes of $50$, the
ratio of the labeled samples in closed-set class samples of $0.2$,
and the usage ratio in open-set class samples of $0.1$. The subset consists of
around 12,000 labeled images and around 200,000 unlabeled images.

\noindent
\textbf{Compared methods}. In the experiments, we select
BigGAN~\cite{Brock2018} with DiffAugment~\cite{Zhao2020} as a base
model and carefully integrate (open-set) semi-supervised methods into it. We
compare OSSGAN with BigGAN~\cite{Brock2018}, RandomGAN, SingleGAN, $S^3$GAN~\cite{Lucic2019}, RejectGAN, and OpensetGAN.
Here, DiffAugment~\cite{Zhao2020} is
applied to the compared methods.  We train BigGAN on only the
labeled samples. In other words, the method is trained on clean
datasets.  The other methods and OSSGAN are trained on our constructed open-set dataset, as mentioned above.
RejectGAN and OpensetGAN are introduced in the above section.
RandomGAN does not have a classifier for the unlabeled images
and assigns labels $y \in \mathcal{Y}$ chosen uniformly at random to unlabeled samples.
RandomGAN is very simple, but it learns reasonably well
when occasionally assigning correct labels by happenstance.
SingleGAN is a naive version of OSSGAN and assigns the high entropy labels to all unlabeled samples regardless of their content.

For OSSGAN, $S^3$GAN, RejectGAN, and OpensetGAN, the weighting parameter
$\lambda$ is selected from $\{0.1, 0.2, 0.4, 0.6\}$, respectively. 
The cross-entropy loss can be large, so care should be taken
to set it such that the training balance between the generator and the
discriminator is not disturbed.  For RejectGAN and OpensetGAN, the threshold $c$ is selected from $\{0.1, 0.3, 0.5, 0.7, 0.9, 0.95\}$.

\noindent
\textbf{Evaluation metrics}.  We employ Inception Score
(IS)~\cite{Salimans2016}, Fr\'{e}chet Inception Distance
(FID)~\cite{Heusel2017}, $F_{1/8}$ score~\cite{Sajjadi2018}, and $F_8$ score~\cite{Sajjadi2018} to measure
the whole quality of generated samples.  FID measures both image
quality and diversity with the feature distance between the generated
and reference images, but it was not possible to separate
the evaluated values into fidelity and diversity. In contrast,
$F_{1/8}$ and $F_8$ aim to quantify fidelity and diversity, respectively.
We sample 10K generated images for all the metrics and use the
evaluation set as the reference distribution for FID.

\section{Experiment results}
\label{sec:result}

\begin{figure}[tb]
\centering
\includegraphics[width=1.0\linewidth]{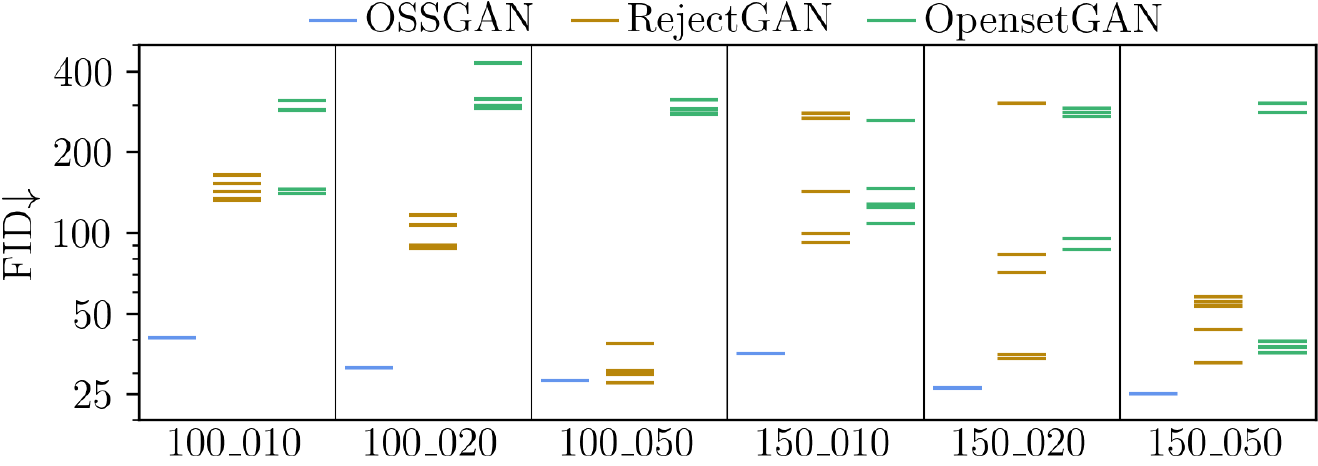}
\caption{Effect of the threshold for detecting open-set class on Tiny ImageNet.
  We only report the
  constant scores for OSSGAN, as it does not employ a threshold. For OpensetGAN and RejectGAN, the optimal
  thresholds are investigated in the range from $0$ to $1$. Our method
  achieves a lower (better) FID than the threshold-based methods because
  of the difficulty of selecting a threshold. The data configuration of $100\_010$ indicates 100 closed classes and 10\% labeled samples. The same notations are applicable to other data configurations.
}\label{fig:compare_threshold}
\end{figure}

 \begin{figure}[tb]
  \centering
    \bgroup 
    \def\arraystretch{0.2} 
    \setlength\tabcolsep{0.2pt}
    \begin{tabular}{cccccc}
\includegraphics[width=0.195\linewidth]{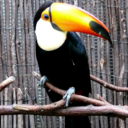} &
\includegraphics[width=0.195\linewidth]{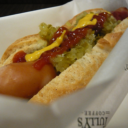} &
\includegraphics[width=0.195\linewidth]{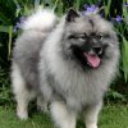} &
\includegraphics[width=0.195\linewidth]{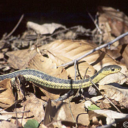} &
\includegraphics[width=0.195\linewidth]{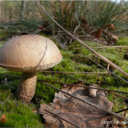} \\
\multicolumn{5}{c}{References} \\ \\
\includegraphics[width=0.195\linewidth]{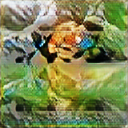} &
\includegraphics[width=0.195\linewidth]{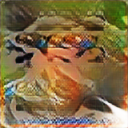} &
\includegraphics[width=0.195\linewidth]{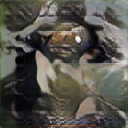} &
\includegraphics[width=0.195\linewidth]{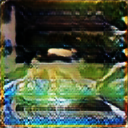} &
\includegraphics[width=0.195\linewidth]{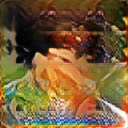} \\ 
\multicolumn{5}{c}{BigGAN~\cite{Brock2018}} \\ \\
\includegraphics[width=0.195\linewidth]{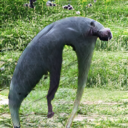} &
\includegraphics[width=0.195\linewidth]{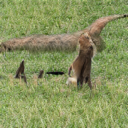} &
\includegraphics[width=0.195\linewidth]{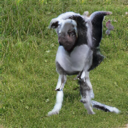} &
\includegraphics[width=0.195\linewidth]{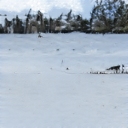} &
\includegraphics[width=0.195\linewidth]{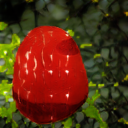} \\ 
\multicolumn{5}{c}{RandomGAN} \\ \\
\includegraphics[width=0.195\linewidth]{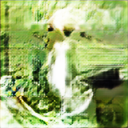} &
\includegraphics[width=0.195\linewidth]{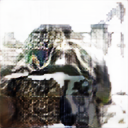} &
\includegraphics[width=0.195\linewidth]{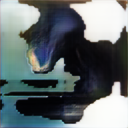} &
\includegraphics[width=0.195\linewidth]{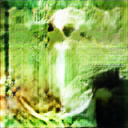} &
\includegraphics[width=0.195\linewidth]{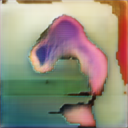} \\ 
\multicolumn{5}{c}{$S^3$GAN~\cite{Lucic2019}} \\ \\
\includegraphics[width=0.195\linewidth]{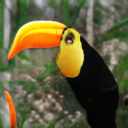} &
\includegraphics[width=0.195\linewidth]{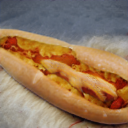} &
\includegraphics[width=0.195\linewidth]{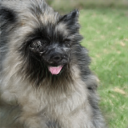} &
\includegraphics[width=0.195\linewidth]{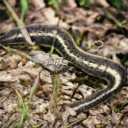} &
\includegraphics[width=0.195\linewidth]{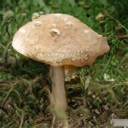} \\ 
\multicolumn{5}{c}{OSSGAN}
    \end{tabular}\egroup
    \caption{Visual comparison of class-conditional image synthesis results on ImageNet.
      Our method produces plausible images while respecting the given condition.
    }\label{fig:generated_examples} 
\end{figure}

\begin{figure}[tb]
\centering
\includegraphics[width=1.0\linewidth]{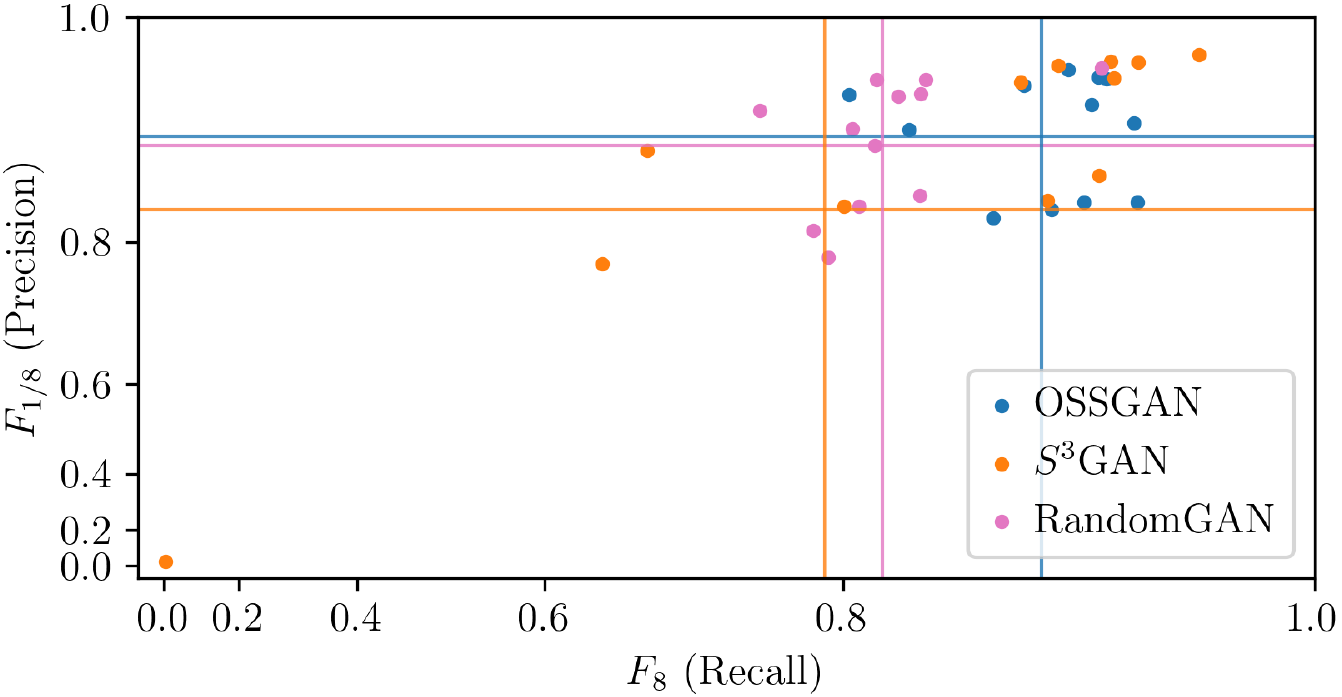}
\caption{Scatter of precision and recall scores obtained in Tiny ImageNet experiments. Dots show
  the recall and precision in experiments.  Horizontal line shows
  the average precision of a method.  Vertical line shows the
  average recall of a method.  Dots in the upper right indicate
  high fidelity and high diversity. Conversely, dots in the
  lower right indicate low fidelity and low
  diversity.
  }\label{fig:fscore}
\end{figure}

\begin{table*}
\centering
\resizebox{2\columnwidth}{!}{\begin{tabular}{cccccccccccccc} \toprule
  \multirow{2}{*}{\shortstack{\#\! known \\ class}} & \multirow{2}{*}{\shortstack{\#\! labeled \\ sample/class}} & \multicolumn{2}{c}{\#\! unlabeled sample}&
  \multicolumn{2}{c}{OSSGAN} & \multicolumn{2}{c}{$S^3$GAN~\cite{Lucic2019}} &
  \multicolumn{2}{c}{RandomGAN} & \multicolumn{2}{c}{SingleGAN} & \multicolumn{2}{c}{BigGAN~\cite{Zhao2020}} \\
  \cmidrule(lr){3-4} \cmidrule(lr){5-6} \cmidrule(lr){7-8} \cmidrule(lr){9-10} \cmidrule(lr){11-12} \cmidrule(lr){13-14}
 &  & closed-set & open-set & FID$\downarrow$ & IS$\uparrow$ & FID$\downarrow$ & IS$\uparrow$ & FID$\downarrow$ & IS$\uparrow$ & FID$\downarrow$ & IS$\uparrow$ & FID$\downarrow$ & IS$\uparrow$ \\ \midrule
\multirow{4}{*}{150}
& 475 & 3750 & 25000 & 25.97 & 15.30 & \phantom{0}20.95 & 16.51 & 26.51 & 12.80 & 24.38 & 13.82 &  $\textbf{\phantom{0}17.82}$ & $\textbf{17.65}$\\
& 250 & 37500 & 25000 & 25.06 & 13.81 & \phantom{0}24.05 & 14.38 & 33.56 & 10.86 & 30.11 & 11.89 &  $\textbf{\phantom{0}18.57}$ & $\textbf{16.99}$\\
& 100 & 60000 & 25000 & $\textbf{26.34}$ & $\textbf{13.66}$ &  \phantom{0}27.25 & 13.12 & 34.13 & 10.82 & 39.67 & 10.81 &  \phantom{0}62.77 &  \phantom{0}9.74\\
& 50 & 67500 & 25000 & $\textbf{35.39}$ & $\textbf{11.61}$ &  \phantom{0}54.32 & \phantom{0}9.36 & 45.23 & 10.78 & 64.25 & 9.68 & 109.50 &  \phantom{0}5.94\\
\midrule
\multirow{4}{*}{100}
& 475 & 2500 & 50000 & 31.75 &13.81 & \phantom{0}28.66 & 14.80  & 36.70 & 11.77& $\textbf{18.66}$ & $\textbf{17.67}$  & $\phantom{0}18.88$ & 17.60\\
& 250 & 25000 & 50000 & 28.19 & 13.88 &\phantom{0}29.19 & 14.05 & 51.33 & 10.02 & 25.97 & 16.66 &  $\textbf{\phantom{0}22.85}$ & $\textbf{17.02}$\\
& 100 & 40000 & 50000 & $\textbf{31.28}$&$\textbf{13.29}$ & \phantom{0}33.18 & 12.60 & 56.50 &  \phantom{0}9.94 & 37.79 & 11.46 &  \phantom{0}56.36 &  \phantom{0}9.82\\
& 50 & 45000 & 50000 & $\textbf{40.61}$& $\textbf{11.54}$ & 230.84 & \phantom{0}2.50 & 42.05 & 10.58 & 43.45 & 10.51 & 128.33 &  \phantom{0}4.59\\
\midrule
\multirow{4}{*}{50}
& 475 & 1250 & 75000 &  56.33 &  13.63 & \phantom{0}55.12 & 13.22 & 68.67 & 10.25 & 71.19 & 9.72 &  $\textbf{\phantom{0}23.81}$ & $\textbf{13.85}$\\
& 250 & 12500 & 75000 & $\textbf{56.49}$ & $\textbf{13.48}$ & \phantom{0}57.16 & 12.50 & 67.98 &  \phantom{0}9.88 & 74.66 & 9.36 &  \phantom{0}78.39 &  \phantom{0}6.47\\
& 100 & 20000 & 75000 & $\textbf{58.36}$ & $\textbf{12.41}$ &  \phantom{0}75.94 & \phantom{0}8.76 & 72.97 &  \phantom{0}9.82 & 77.17 & 9.14 &  \phantom{0}99.78 &  \phantom{0}5.14\\
& 50 & 22500 & 75000 & $\textbf{61.60}$& $\textbf{11.84}$ & \phantom{0}95.01 & \phantom{0}8.29 & 79.75 &  \phantom{0}8.53 & 73.41 & 9.08 & 161.65 &  \phantom{0}4.17\\
\bottomrule
\end{tabular}}
\caption{FID and Inception Scores obtained in experiments of different
  configurations on Tiny ImageNet. Each row shows the configuration, statistics, and
  results of compared methods in an experiment. 
  }\label{fig:fid_tiny}
\end{table*}

\begin{table}
\centering
\begin{tabular}{lcccc} \toprule
                           & FID$\downarrow$                & IS$\uparrow$                & $F_{1/8}$$\uparrow$          & $F_{8}$$\uparrow$          \\ \midrule
  \!\!BigGAN~\cite{Zhao2020}  & 190.88           & \phantom{0}3.97             & 0.1178            & 0.0570           \\
  \!\!RandomGAN               & 105.71           & 12.41            & 0.6104            & 0.7679           \\
  \!\!$S^3$GAN~\cite{Lucic2019}  & 180.30 & \phantom{0}4.38  & 0.2053 & 0.1472 \\
  \!\!OSSGAN                   & \phantom{0}$\textbf{78.43}$ & $\textbf{18.42}$ & $\textbf{0.8379}$ & $\textbf{0.8359}$\\
  \bottomrule
\end{tabular}
\caption{Experimental results for our ImageNet dataset. 
Our method outperforms the baseline methods in terms of all the quantitative metrics.
}\label{tb:imagenet}
\end{table}

\begin{table}
\centering
\begin{tabular}{lcc} \toprule
 & Tiny\! ImageNet & ImageNet \\ \midrule
Ours w/o $H$ and Fake  & 60.09 & 117.78\\
Ours w/o $H$ & 70.55 & \phantom{0}85.65\\
Ours w/o Fake  & 60.06 & \phantom{0}81.53\\
Ours (OSSGAN) & \textbf{35.39} & \phantom{0}\textbf{78.43}\\
  \bottomrule
\end{tabular}
\caption{Results of ablation study. We report the FID scores of each method. 
In the ablation parts, $H$ and Fake indicate the entropy regularization term and the cross-entropy loss for fake samples, respectively.
The last row is for our OSSGAN model.}\label{tb:ablation_study}
\end{table}

We first evaluate the effects of the entropy regularization term.
To this end, we compute the difference in average entropies at the 100k-th iteration.
The difference without the regularization term is 0.104, whereas the difference with the regularization term is 0.283.
This shows that optimizing a classifier with entropy regularization results in clearly separated open-set and closed-set samples in unlabeled samples.
Consequently, we can expect the regularization to improve the cGAN model performance even when the model is trained on a dataset contaminated with open-set samples.

We then investigate whether incorporating the threshold has a negative impact on cGAN training. The FID scores in baseline methods with different thresholds and in OSSGAN are shown in \cref{fig:compare_threshold}.
We see that the quantitative performances of OpensetGAN and RejectGAN are heavily dependent on threshold selection.
Their results are not on par with our results, except in the best-case scenario.
This is because the average entropy of the open-set samples varies greatly during training, ranging from $0.4$ to $0.9$. 
We obtain the average entropies of open-set and closed-set samples, only for benchmarking purposes, not in practice, because we cannot divide unlabeled samples into open-set and closed-set samples. 
Because we cannot set a threshold with respect to the average entropy, the method free of threshold adjustments is more useful for our purposes.

We next compare OSSGAN to $S^3$GAN, RandomGAN, SingleGAN, and BigGAN on 12 data configurations of Tiny ImageNet, as decribed in \cref{sec:experiment}.
\Cref{fig:fid_tiny} shows the quantitative results separated into three segments.
Each segment has the same number of closed-set classes while it has different ratios of labeled samples in the closed-set classes.
The total number of samples (labeled and unlabeled samples) in each experiment is 100,000.
We note that the lower the row, the more difficult the experiment.

We summarize the experimental results of our Tiny ImageNet datasets.
For the dataset with sufficient labeled samples for each class (\eg, rows 1, 2, and 5), it is not surprising that BigGAN performs best thanks to the use of data-efficient DiffAugment module.
In contrast, when the data difficulty increases, its performance degrades drastically.
RandomGAN and SingleGAN perform better than BigGAN for datasets with insufficient number of labeled samples for each class.
While $S^3$GAN provides further improvements over the methods that do not rely on the auxiliary classifier for such datasets, it worsens in extreme cases.
Our method outperforms the baselines for the dataset with limited number of labeled samples even in extreme cases.
The performance gains are a result of smoothly utilizing
the likelihoods quantified by the classifier. OSSGAN with 20\%
labeled samples achieves comparable performance to BigGAN (see row 10 and row 12 of \cref{fig:fid_tiny}). 

We also conduct experiments with high-resolution images for a subset
of ImageNet. In this experiment, we use labeled data from only 1\% of the
original ImageNet.  \Cref{tb:imagenet} shows the $F_{1/8}$, $F_8$,
FID, and IS scores for the dataset mentioned in \cref{sec:experiment}. BigGAN and $S^3$GAN completely fail for this
dataset.  The low $F_{1/8}$ and high $F_8$ scores indicate that RandomGAN
probably fails to generate images of a class that matches the given
class condition.  Our method achieves the best quantitative performance in
all metrics.  \Cref{fig:generated_examples} also shows the examples
generated by the methods. The qualitative results are consistent
with the quantitative results.

The $F_{1/8}$ and $F_8$ scores of compared methods in Tiny ImageNet
experiments with several configurations and the average $F_{1/8}$ and
$F_8$ scores of each method are shown in \cref{fig:fscore}.  The proposed method has a higher recall,
that is, diversity, than other methods.
RandomGAN, which assigns all classes to each image, fails to learn class-specific features, resulting in low diversity.
In contrast, our method learns
class-specific and class-invariant features by utilizing confidence,
leading to achieve high diversity.

We perform an ablation study to evaluate the contribution of each loss term of our model. 
For ablation models, we drop either or both the entropy regularization term and the cross-entropy loss for fake samples.
The components are indicated by $H$ and Fake, respectively.
As \cref{tb:ablation_study} shows, both components contribute to the performance individually, and the combination of the components yields significant improvement.
Applying only Fake sometimes harms the performance in the case of a sufficient number of labeled samples for training the classifier.

\section{Conclusion}
\label{sec:conclusion}

We introduced a novel task of practical image generation.  We
proposed open-set semi-supervised image generation, a problem that
takes into account the properties of the data available when
developing applications for image generation models in the real world.
Furthermore, to address the proposed task, we designed OSSGAN by integrating the unlabeled samples with the confidence score obtained by the auxiliary classifier into the training of cGANs. Thanks to the utilization of entropy regularization, OSSGAN promoted the discriminator to learn the class-invariant features and to avoid missing the useful features of closed-set classes.
The results of our comprehensive
experiments on several configurations showed that OSSGAN outperformed other
baseline methods and performed well with limited labeled samples. 
Therefore, our OSSGAN will reduce the cost of building datasets for training cGANs,  leading to the expansion of the range of real-world applications of cGANs.
The limitation of the proposed method is that the performance improvement by OSSGAN depends on
the success of the training of the classifier. In combination with
unsupervised and self-supervised learning, we expect to be able to
achieve learning of appropriate classifiers from fewer labels than in our method.

\noindent
\textbf{Acknowledgement.}
This work was supported by D-CORE Grant from Microsoft Research Asia, the Institute of AI and Beyond of the University of Tokyo, the Next Generation Artificial Intelligence Research Center of the University of Tokyo, and JSPS KAKENHI Grant Number JP19H04166.

{\small \bibliographystyle{ieee_fullname} \bibliography{submission} }

\clearpage

\appendix

\setcounter{figure}{0}

\renewcommand\thesection{\Roman{section}}
\renewcommand\thesubsection{\Roman{section}.\Alph{subsection}}
\renewcommand\thefigure{\Alph{figure}} 
\renewcommand\thetable{\Alph{table}} 

\twocolumn[\centering{\Large \bf Supplementary Material for \\ OSSGAN: Open-Set Semi-Supervised Image Generation \\ \vspace*{\baselineskip}}]

\justifying

\section{Algorithm of the proposed method}
\Cref{alg:osssgan} shows the algorithm of Softlabel-GAN. The method
has a few modifications from supervised GANs, resulting in easily
applying to other cGAN architectures instead of BigGAN.
\begin{algorithm}[ht]
\DontPrintSemicolon
\KwData{Generator $G$, Discriminator $D$, Classifier $C$, labeled data $\mathcal{D}_t$, unlabeled data $\mathcal{D}_u$, total number of iteration $T$}
\KwResult{Trained $G$ and $D$}
   initialize $\{\Gparam, \Dparam, \Cparam\}$\;
   \For{$t \gets 1$ \KwTo $T$}{
     Sample Batch $(\x, \y) \sim \mathcal{D}_t, \ux \sim \mathcal{D}_u, (\z, \y^{f})\sim q(\z, \y)$\;  
     Calculate $\mathcal{L}_{D}$ using Eq. 9\;
     $\Dparam, \Cparam \gets \Adam(\mathcal{L}_{D}, \{\Dparam, \Cparam\})$\;
     \;
     Sample Batch $\qzy$\;
     Calculate $\mathcal{L}_G$ using Eq. 10\;
     $\Gparam \gets \Adam(\mathcal{L}_{G}, \Gparam)$\;
    }
\caption{Open-Set Semi-Supervised GAN.}\label{alg:osssgan}
\end{algorithm}

\section{Intuitive illustration}

\begin{figure}[tb]
\centering
\includegraphics[width=1.0\linewidth]{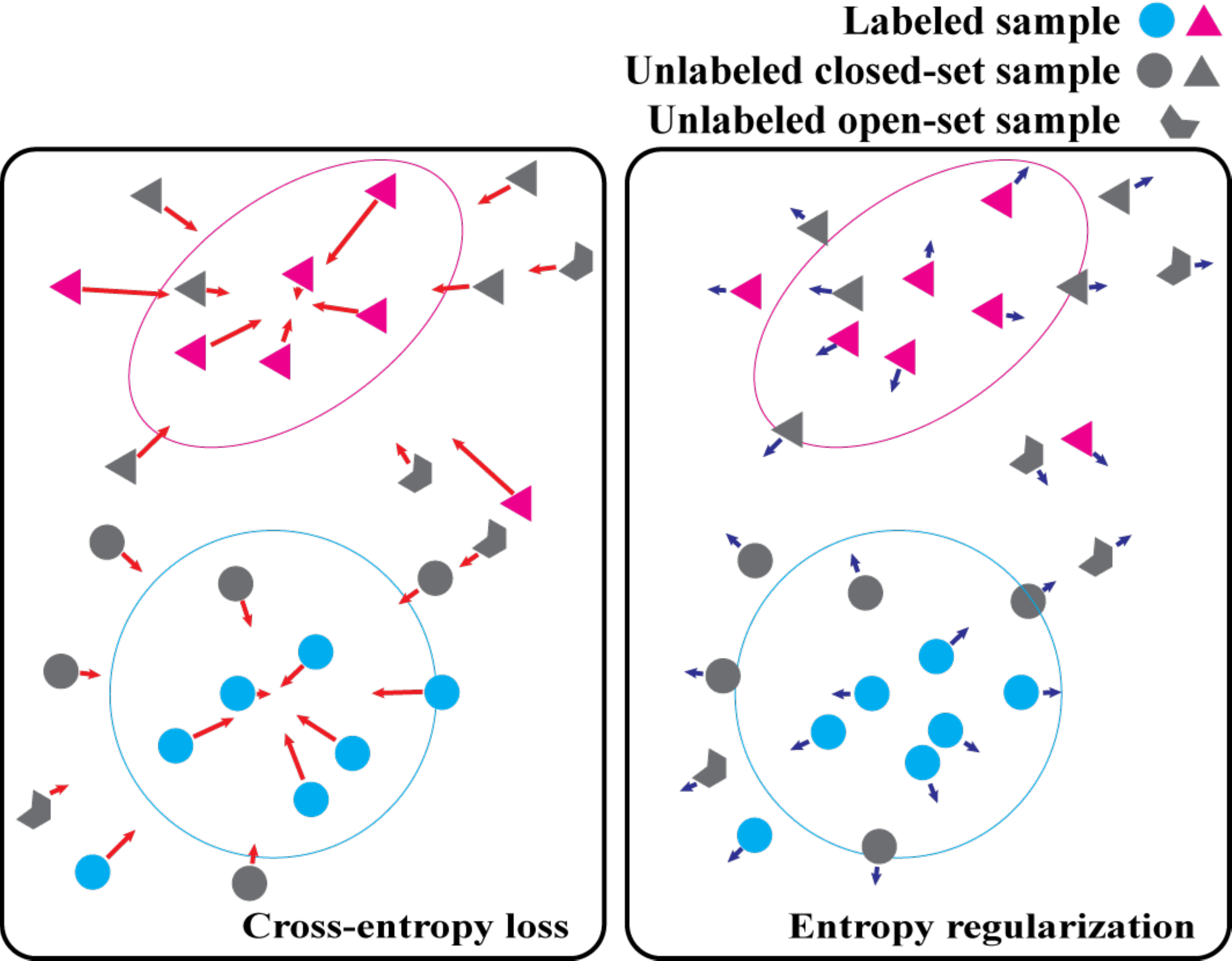}
\caption{Intuitive illustration about how does our method work.}\label{fig:method_behavior}
\end{figure}

As \cref{fig:method_behavior} shows, while the cross-entropy loss
makes the entropy smaller, the entropy regularization makes the
entropy larger. The cross-entropy loss affects close-set samples
stronger, resulting in the clear separation between the closed- and
open-set samples.

\section{Comparison with additional threshold-based methods}

\begin{figure}[tb]
\centering
\includegraphics[width=1.0\linewidth]{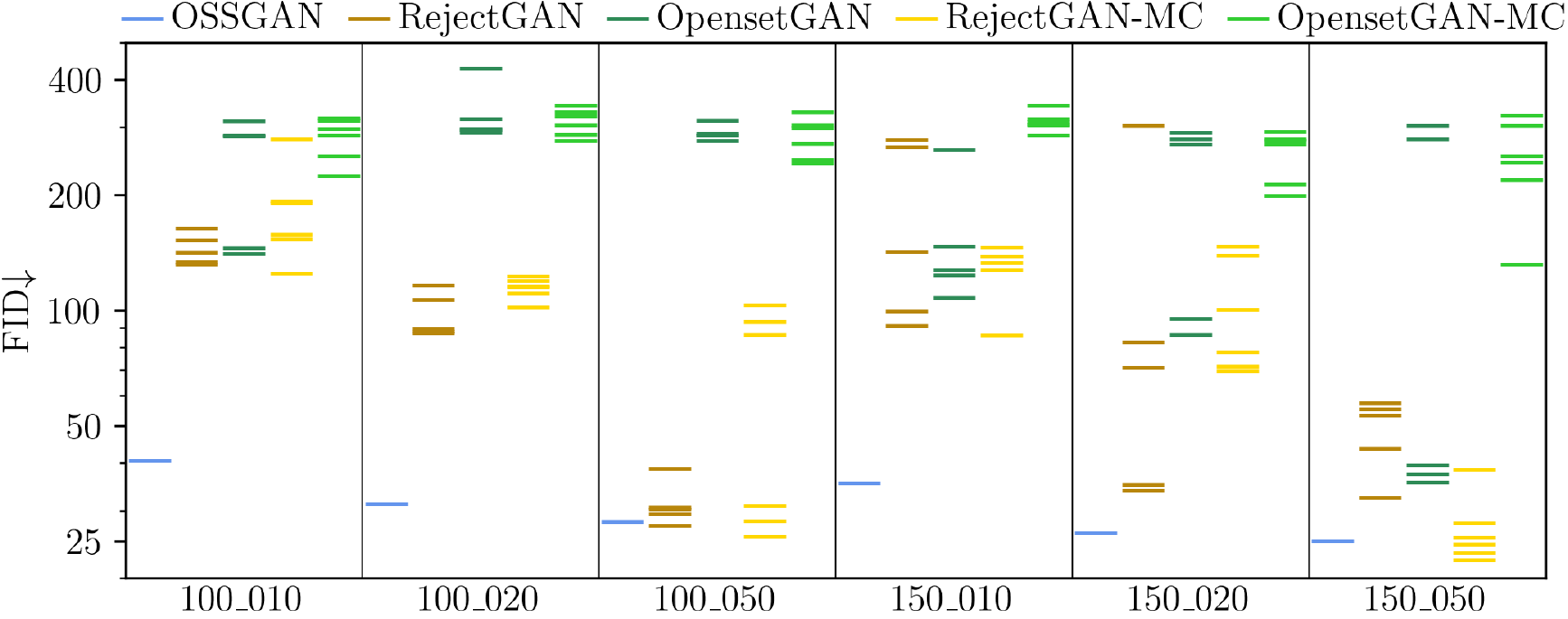}
\caption{Comparison of OSSGAN and threshold-based methods.  We only
  report the constant scores for OSSGAN, as it does not employ a
  threshold. For OpensetGAN and RejectGAN, the optimal thresholds are
  investigated in the range from $0$ to $1$.  The range for
  OpensetGAN-MC and RejectGAN-MC is from $0.000001$ to $0.0001$ 
  The threshold-based methods fail in most cases
  due to the difficulty of selecting a threshold. The data
  configuration of $100\_010$ indicates 100 closed classes and 10\%
  labeled samples. The same notations are applicable to other data
  configurations.  }\label{fig:compare_threshold_mc}
\end{figure}

In addition to comparison with threshold-based methods with ad-hoc labeling schemes in Sec. 6, we further compare our OSSGAN with threshold-based methods with Montre-Carlo Dropout
uncertainty estimation in \cref{fig:compare_threshold_mc}. 
OpensetGAN-MC and RejectGAN-MC indicate OpensetGAN with MC Dropout and RejectGAN with MC Dropout, respectively.
As \cref{fig:compare_threshold_mc} shows, RejectGAN-MC performs on par with OSSGAN in only a few cases with the easy configuration and the best threshold.
In other cases, it is still difficult to achieve better performance for threshold-based methods with MC Dropout.
These results show that threshold-based methods can not work for our complex task regardless of the quality of quantified confidence.

\section{More examples}
\begin{figure}[tb]
  \centering
    \bgroup 
    \def\arraystretch{0.2} 
    \setlength\tabcolsep{0.2pt}
    \begin{tabular}{cccccc}
\includegraphics[width=0.195\linewidth,height=0.195\linewidth]{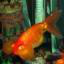} &
\includegraphics[width=0.195\linewidth,height=0.195\linewidth]{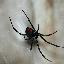} &
\includegraphics[width=0.195\linewidth,height=0.195\linewidth]{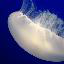} &
\includegraphics[width=0.195\linewidth,height=0.195\linewidth]{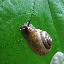} &
\includegraphics[width=0.195\linewidth,height=0.195\linewidth]{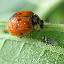} \\
\multicolumn{5}{c}{References} \\ \\
\includegraphics[width=0.195\linewidth,height=0.195\linewidth]{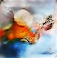} &
\includegraphics[width=0.195\linewidth,height=0.195\linewidth]{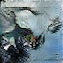} &
\includegraphics[width=0.195\linewidth,height=0.195\linewidth]{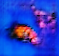} &
\includegraphics[width=0.195\linewidth,height=0.195\linewidth]{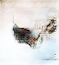} &
\includegraphics[width=0.195\linewidth,height=0.195\linewidth]{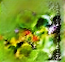} \\ 
\multicolumn{5}{c}{BigGAN} \\ \\
\includegraphics[width=0.195\linewidth,height=0.195\linewidth]{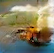} &
\includegraphics[width=0.195\linewidth,height=0.195\linewidth]{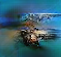} &
\includegraphics[width=0.195\linewidth,height=0.195\linewidth]{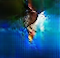} &
\includegraphics[width=0.195\linewidth,height=0.195\linewidth]{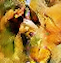} &
\includegraphics[width=0.195\linewidth,height=0.195\linewidth]{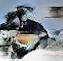} \\ 
\multicolumn{5}{c}{RandomGAN} \\ \\
\includegraphics[width=0.195\linewidth,height=0.195\linewidth]{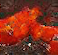} &
\includegraphics[width=0.195\linewidth,height=0.195\linewidth]{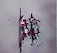} &
\includegraphics[width=0.195\linewidth,height=0.195\linewidth]{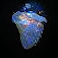} &
\includegraphics[width=0.195\linewidth,height=0.195\linewidth]{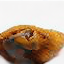} &
\includegraphics[width=0.195\linewidth,height=0.195\linewidth]{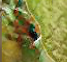} \\
\multicolumn{5}{c}{$S^3$GAN} \\ \\
\includegraphics[width=0.195\linewidth,height=0.195\linewidth]{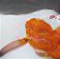} &
\includegraphics[width=0.195\linewidth,height=0.195\linewidth]{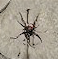} &
\includegraphics[width=0.195\linewidth,height=0.195\linewidth]{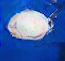} &
\includegraphics[width=0.195\linewidth,height=0.195\linewidth]{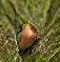} &
\includegraphics[width=0.195\linewidth,height=0.195\linewidth]{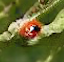} \\
\multicolumn{5}{c}{OSSGAN}
    \end{tabular}\egroup
    \caption{Visual comparison of class-conditional image synthesis
      results on Tiny ImageNet with 50 classes. Our method produces
      plausible images while respecting the given
      condition.}\label{fig:generated_tiny}
\end{figure}
\begin{figure}[tb]
  \centering
    \bgroup 
    \def\arraystretch{0.2} 
    \setlength\tabcolsep{0.2pt}
    \begin{tabular}{ccccc}
\includegraphics[width=0.195\linewidth]{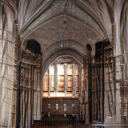} &
\includegraphics[width=0.195\linewidth]{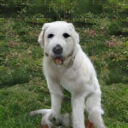} &
\includegraphics[width=0.195\linewidth]{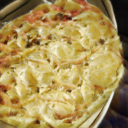} &
\includegraphics[width=0.195\linewidth]{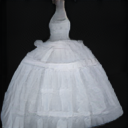} &
\includegraphics[width=0.195\linewidth]{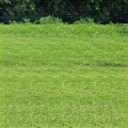} \\
\includegraphics[width=0.195\linewidth]{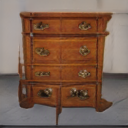} &
\includegraphics[width=0.195\linewidth]{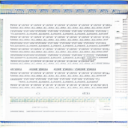} &
\includegraphics[width=0.195\linewidth]{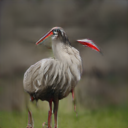} &
\includegraphics[width=0.195\linewidth]{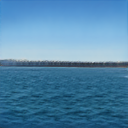} &
\includegraphics[width=0.195\linewidth]{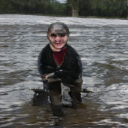} \\
\includegraphics[width=0.195\linewidth]{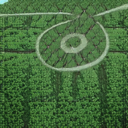} &
\includegraphics[width=0.195\linewidth]{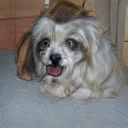} &
\includegraphics[width=0.195\linewidth]{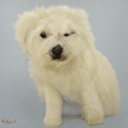} &
\includegraphics[width=0.195\linewidth]{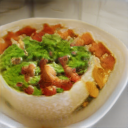} &
\includegraphics[width=0.195\linewidth]{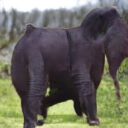} \\
\includegraphics[width=0.195\linewidth]{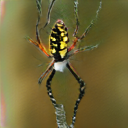} &
\includegraphics[width=0.195\linewidth]{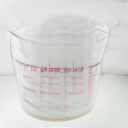} &
\includegraphics[width=0.195\linewidth]{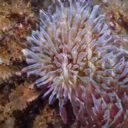} &
\includegraphics[width=0.195\linewidth]{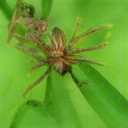} &
\includegraphics[width=0.195\linewidth]{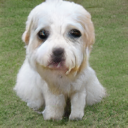} \\
\includegraphics[width=0.195\linewidth]{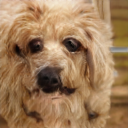} &
\includegraphics[width=0.195\linewidth]{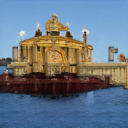} &
\includegraphics[width=0.195\linewidth]{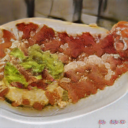} &
\includegraphics[width=0.195\linewidth]{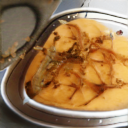} &
\includegraphics[width=0.195\linewidth]{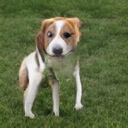} \\
\includegraphics[width=0.195\linewidth]{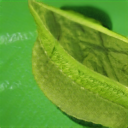} &
\includegraphics[width=0.195\linewidth]{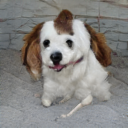} &
\includegraphics[width=0.195\linewidth]{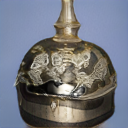} &
\includegraphics[width=0.195\linewidth]{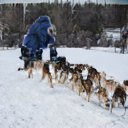} &
\includegraphics[width=0.195\linewidth]{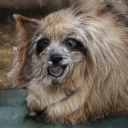} 
    \end{tabular}\egroup
    \caption{More examples synthesized by OSSGAN on ImageNet with 50 classes.}\label{fig:generated_imagenet}
\end{figure}
\Cref{fig:generated_tiny} provides generated examples from the
compared methods. Our method produces plausible images while the other
methods fail to produce plausible images.
$S^3$GAN produces images respecting the given condition but lacking the plausibility of images.

We provide more generated examples of OSSGAN on ImageNet with 50
classes in \cref{fig:generated_imagenet}.

\begin{figure*}[tb]
  \begin{minipage}{1\columnwidth}
  \centering
    \bgroup 
    \def\arraystretch{0.2} 
    \setlength\tabcolsep{0.2pt}
    \begin{tabular}{ccccc}
\includegraphics[width=0.195\linewidth]{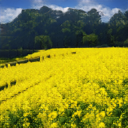} &
\includegraphics[width=0.195\linewidth]{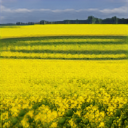} &
\includegraphics[width=0.195\linewidth]{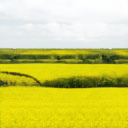} &
\includegraphics[width=0.195\linewidth]{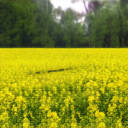} &
\includegraphics[width=0.195\linewidth]{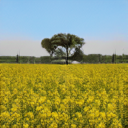} \\
\includegraphics[width=0.195\linewidth]{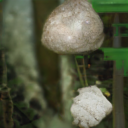} &
\includegraphics[width=0.195\linewidth]{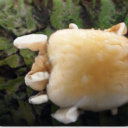} &
\includegraphics[width=0.195\linewidth]{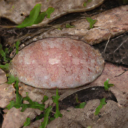} &
\includegraphics[width=0.195\linewidth]{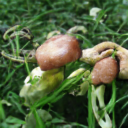} &
\includegraphics[width=0.195\linewidth]{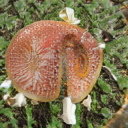} \\
\includegraphics[width=0.195\linewidth]{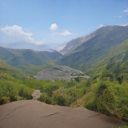} &
\includegraphics[width=0.195\linewidth]{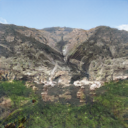} &
\includegraphics[width=0.195\linewidth]{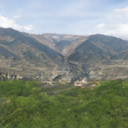} &
\includegraphics[width=0.195\linewidth]{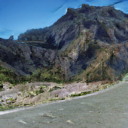} &
\includegraphics[width=0.195\linewidth]{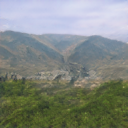} \\
\includegraphics[width=0.195\linewidth]{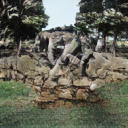} &
\includegraphics[width=0.195\linewidth]{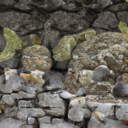} &
\includegraphics[width=0.195\linewidth]{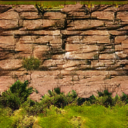} &
\includegraphics[width=0.195\linewidth]{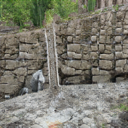} &
\includegraphics[width=0.195\linewidth]{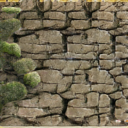} \\
\includegraphics[width=0.195\linewidth]{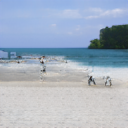} &
\includegraphics[width=0.195\linewidth]{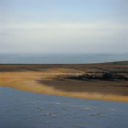} &
\includegraphics[width=0.195\linewidth]{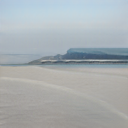} &
\includegraphics[width=0.195\linewidth]{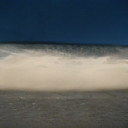} &
\includegraphics[width=0.195\linewidth]{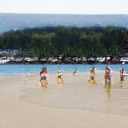} \\
\includegraphics[width=0.195\linewidth]{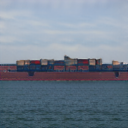} &
\includegraphics[width=0.195\linewidth]{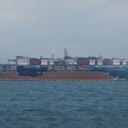} &
\includegraphics[width=0.195\linewidth]{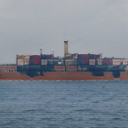} &
\includegraphics[width=0.195\linewidth]{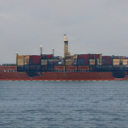} &
\includegraphics[width=0.195\linewidth]{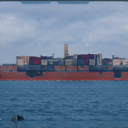} \\
\includegraphics[width=0.195\linewidth]{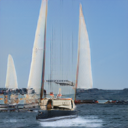} &
\includegraphics[width=0.195\linewidth]{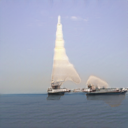} &
\includegraphics[width=0.195\linewidth]{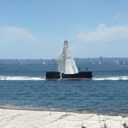} &
\includegraphics[width=0.195\linewidth]{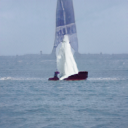} &
\includegraphics[width=0.195\linewidth]{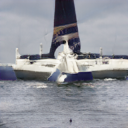} \\
\includegraphics[width=0.195\linewidth]{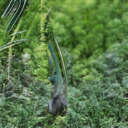} &
\includegraphics[width=0.195\linewidth]{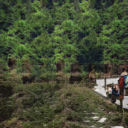} &
\includegraphics[width=0.195\linewidth]{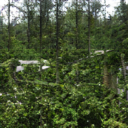} &
\includegraphics[width=0.195\linewidth]{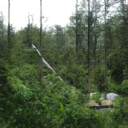} &
\includegraphics[width=0.195\linewidth]{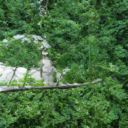} \\
\includegraphics[width=0.195\linewidth]{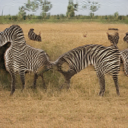} &
\includegraphics[width=0.195\linewidth]{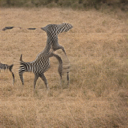} &
\includegraphics[width=0.195\linewidth]{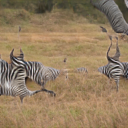} &
\includegraphics[width=0.195\linewidth]{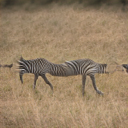} &
\includegraphics[width=0.195\linewidth]{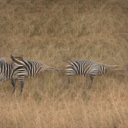} \\
\includegraphics[width=0.195\linewidth]{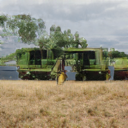} &
\includegraphics[width=0.195\linewidth]{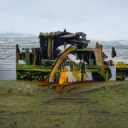} &
\includegraphics[width=0.195\linewidth]{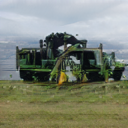} &
\includegraphics[width=0.195\linewidth]{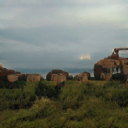} &
\includegraphics[width=0.195\linewidth]{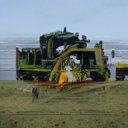} \\
\includegraphics[width=0.195\linewidth]{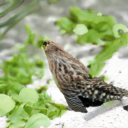} &
\includegraphics[width=0.195\linewidth]{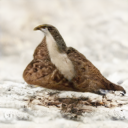} &
\includegraphics[width=0.195\linewidth]{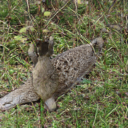} &
\includegraphics[width=0.195\linewidth]{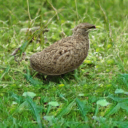} &
\includegraphics[width=0.195\linewidth]{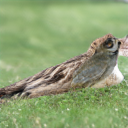} \\
\includegraphics[width=0.195\linewidth]{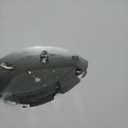} &
\includegraphics[width=0.195\linewidth]{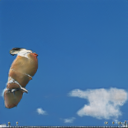} &
\includegraphics[width=0.195\linewidth]{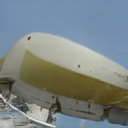} &
\includegraphics[width=0.195\linewidth]{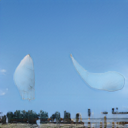} &
\includegraphics[width=0.195\linewidth]{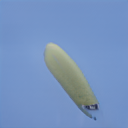} 
    \end{tabular}\egroup
    \subcaption{$S^3$GAN}
  \end{minipage}\hfill
  \begin{minipage}{1\columnwidth}
  \centering
    \bgroup 
    \def\arraystretch{0.2} 
    \setlength\tabcolsep{0.2pt}
    \begin{tabular}{ccccc}
\includegraphics[width=0.195\linewidth]{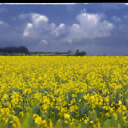} &
\includegraphics[width=0.195\linewidth]{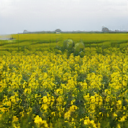} &
\includegraphics[width=0.195\linewidth]{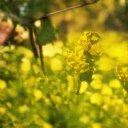} &
\includegraphics[width=0.195\linewidth]{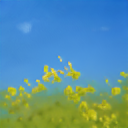} &
\includegraphics[width=0.195\linewidth]{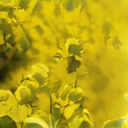} \\
\includegraphics[width=0.195\linewidth]{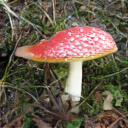} &
\includegraphics[width=0.195\linewidth]{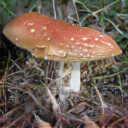} &
\includegraphics[width=0.195\linewidth]{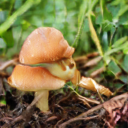} &
\includegraphics[width=0.195\linewidth]{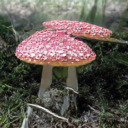} &
\includegraphics[width=0.195\linewidth]{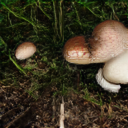} \\
\includegraphics[width=0.195\linewidth]{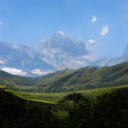} &
\includegraphics[width=0.195\linewidth]{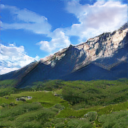} &
\includegraphics[width=0.195\linewidth]{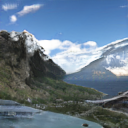} &
\includegraphics[width=0.195\linewidth]{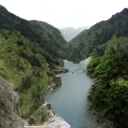} &
\includegraphics[width=0.195\linewidth]{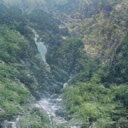} \\
\includegraphics[width=0.195\linewidth]{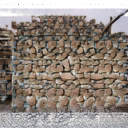} &
\includegraphics[width=0.195\linewidth]{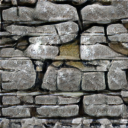} &
\includegraphics[width=0.195\linewidth]{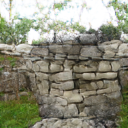} &
\includegraphics[width=0.195\linewidth]{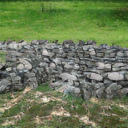} &
\includegraphics[width=0.195\linewidth]{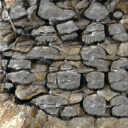} \\
\includegraphics[width=0.195\linewidth]{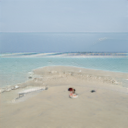} &
\includegraphics[width=0.195\linewidth]{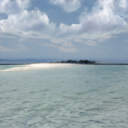} &
\includegraphics[width=0.195\linewidth]{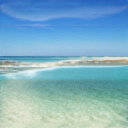} &
\includegraphics[width=0.195\linewidth]{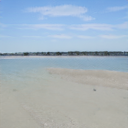} &
\includegraphics[width=0.195\linewidth]{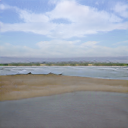} \\
\includegraphics[width=0.195\linewidth]{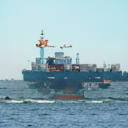} &
\includegraphics[width=0.195\linewidth]{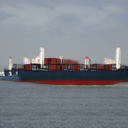} &
\includegraphics[width=0.195\linewidth]{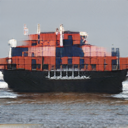} &
\includegraphics[width=0.195\linewidth]{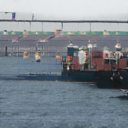} &
\includegraphics[width=0.195\linewidth]{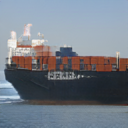} \\
\includegraphics[width=0.195\linewidth]{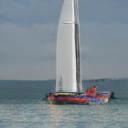} &
\includegraphics[width=0.195\linewidth]{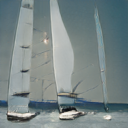} &
\includegraphics[width=0.195\linewidth]{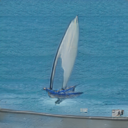} &
\includegraphics[width=0.195\linewidth]{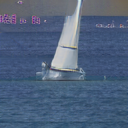} &
\includegraphics[width=0.195\linewidth]{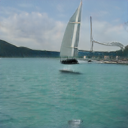} \\
\includegraphics[width=0.195\linewidth]{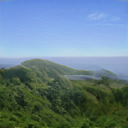} &
\includegraphics[width=0.195\linewidth]{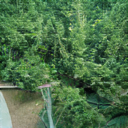} &
\includegraphics[width=0.195\linewidth]{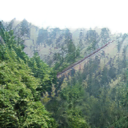} &
\includegraphics[width=0.195\linewidth]{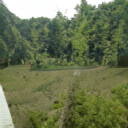} &
\includegraphics[width=0.195\linewidth]{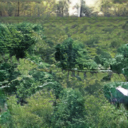} \\
\includegraphics[width=0.195\linewidth]{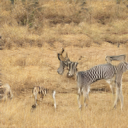} &
\includegraphics[width=0.195\linewidth]{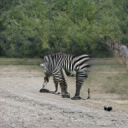} &
\includegraphics[width=0.195\linewidth]{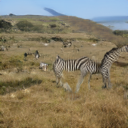} &
\includegraphics[width=0.195\linewidth]{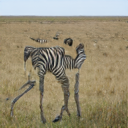} &
\includegraphics[width=0.195\linewidth]{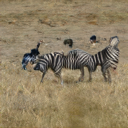} \\
\includegraphics[width=0.195\linewidth]{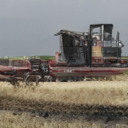} &
\includegraphics[width=0.195\linewidth]{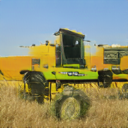} &
\includegraphics[width=0.195\linewidth]{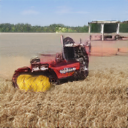} &
\includegraphics[width=0.195\linewidth]{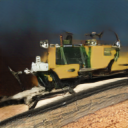} &
\includegraphics[width=0.195\linewidth]{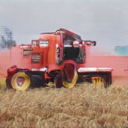} \\
\includegraphics[width=0.195\linewidth]{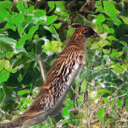} &
\includegraphics[width=0.195\linewidth]{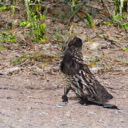} &
\includegraphics[width=0.195\linewidth]{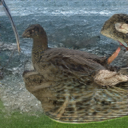} &
\includegraphics[width=0.195\linewidth]{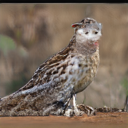} &
\includegraphics[width=0.195\linewidth]{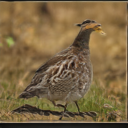} \\
\includegraphics[width=0.195\linewidth]{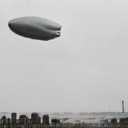} &
\includegraphics[width=0.195\linewidth]{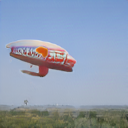} &
\includegraphics[width=0.195\linewidth]{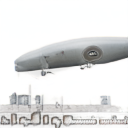} &
\includegraphics[width=0.195\linewidth]{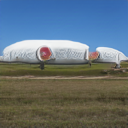} &
\includegraphics[width=0.195\linewidth]{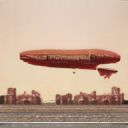} 
    \end{tabular}\egroup
    \subcaption{OSSGAN}
  \end{minipage}
  \caption{Qualitative comparison of class-conditional image synthesis
    results on ImageNet with 100 classes. Our method produces diverse
    images.}\label{fig:imagenet100cls}
\end{figure*}

We also conduct experiments on ImageNet. The experiments have the
number of closed-set classes of $100$, the ratio of the labeled
samples in closed-set class samples of $0.2$, and the usage ratio in
open-set class samples of $0.1$.  Our OSSGAN achieves an IS of $22.11$
and FID of $45.42$, improving over $S^3$GAN with an IS of $18.79$ and
FID of $51.96$, RandomGAN with an IS of $11.04$ and FID of $99.00$,
and BigGAN with an IS of $4.31$ and FID of $182.90$. The qualitative
results of OSSGAN and $S^3$GAN are shown in
\cref{fig:imagenet100cls}. In contrast to $S^3$GAN sometimes generate
While $S^3$GAN sometimes generates almost the same images repeatedly,
OSSGAN generates diverse and plausible images.

\end{document}